\documentclass{article} 
\usepackage{iclr2024_conference,times}

\usepackage{amsmath,amsfonts,bm}

\def\eqref#1{equation~\ref{#1}}

\def\1{\bm{1}}

\DeclareMathAlphabet{\mathsfit}{\encodingdefault}{\sfdefault}{m}{sl}
\SetMathAlphabet{\mathsfit}{bold}{\encodingdefault}{\sfdefault}{bx}{n}

\usepackage{hyperref}
\usepackage{url}
\iclrfinalcopy

\usepackage{xspace} 
\usepackage{graphicx}
\usepackage{booktabs} 
\newcommand{\MathChat}{{\texttt{MathChat}}\xspace}

\usepackage{enumitem}

\title{MathChat: Converse to Tackle Challenging Math Problems with LLM Agents}

\author{Yiran Wu\textsuperscript{1}, Feiran Jia\textsuperscript{1}, Shaokun Zhang\textsuperscript{1}, Hangyu Li\textsuperscript{2}, Erkang Zhu\textsuperscript{3}, Yue Wang\textsuperscript{3},\\
\textbf{Yin Tat Lee\textsuperscript{4}, Richard Peng\textsuperscript{5}, Qingyun Wu\textsuperscript{1}, Chi Wang\textsuperscript{3}}
\\
\textsuperscript{1}Pennsylvania State University \textsuperscript{2}Imperial College London
\textsuperscript{3}Microsoft Research Redmond\\
\textsuperscript{4}University of Washington
\textsuperscript{5}University of Waterloo
\\
\texttt{\{yiran.wu, feiran.jia, shaokun.zhang, qingyun.wu\}@psu.edu},\\
\texttt{\{ekzhu, wang.yue, wang.chi\}@microsoft.com}, \texttt{hl6021@ic.ac.uk},\\ 
\texttt{yintat@uw.edu}, \texttt{y5peng@uwaterloo.ca}
}

\begin{document}

\maketitle

\begin{abstract}
Employing Large Language Models (LLMs) to address mathematical problems is an intriguing research endeavor, considering the abundance of math problems expressed in natural language across numerous science and engineering fields. LLMs, with their generalized ability, are used as a foundation model to build AI agents for different tasks. In this paper, we study the effectiveness of utilizing LLM agents to solve math problems through conversations. We propose \MathChat, a conversational problem-solving framework designed for math problems. \MathChat consists of an LLM agent and a user proxy agent which is responsible for tool execution and additional guidance. This synergy facilitates a collaborative problem-solving process, where the agents engage in a dialogue to solve the problems.
We perform evaluation on difficult high school competition problems from the MATH dataset.  Utilizing Python, we show that \MathChat can further improve previous tool-using prompting methods by 6\%.

\end{abstract}

\vspace{-1em}

\section{Introduction}
\vspace{-0.5em}

With Large Language Models (LLMs) demonstrating remarkable proficiency in various tasks spanning diverse domains~\citep{bubeck2023sparks,zhang2023ideal}, they are deemed the potential foundation model for building autonomous agents~\citep{xi2023rise, wang2023survey, zhang2024training}. Especially, multi-agent collaboration is a promising direction with the growing complexity of tasks being studied, with the benefit of information sharing and collective decision-making among different agents with specialized skills. It is compelling to explore the potential of LLMs in tackling mathematical problems considering the crucial role of mathematics~\citep{wigner1990unreasonable} and the prevalence of mathematical problems expressed in natural language throughout numerous scientific and engineering disciplines. 

\begin{figure}[tb]
\centering
\includegraphics[width = 0.98\hsize]{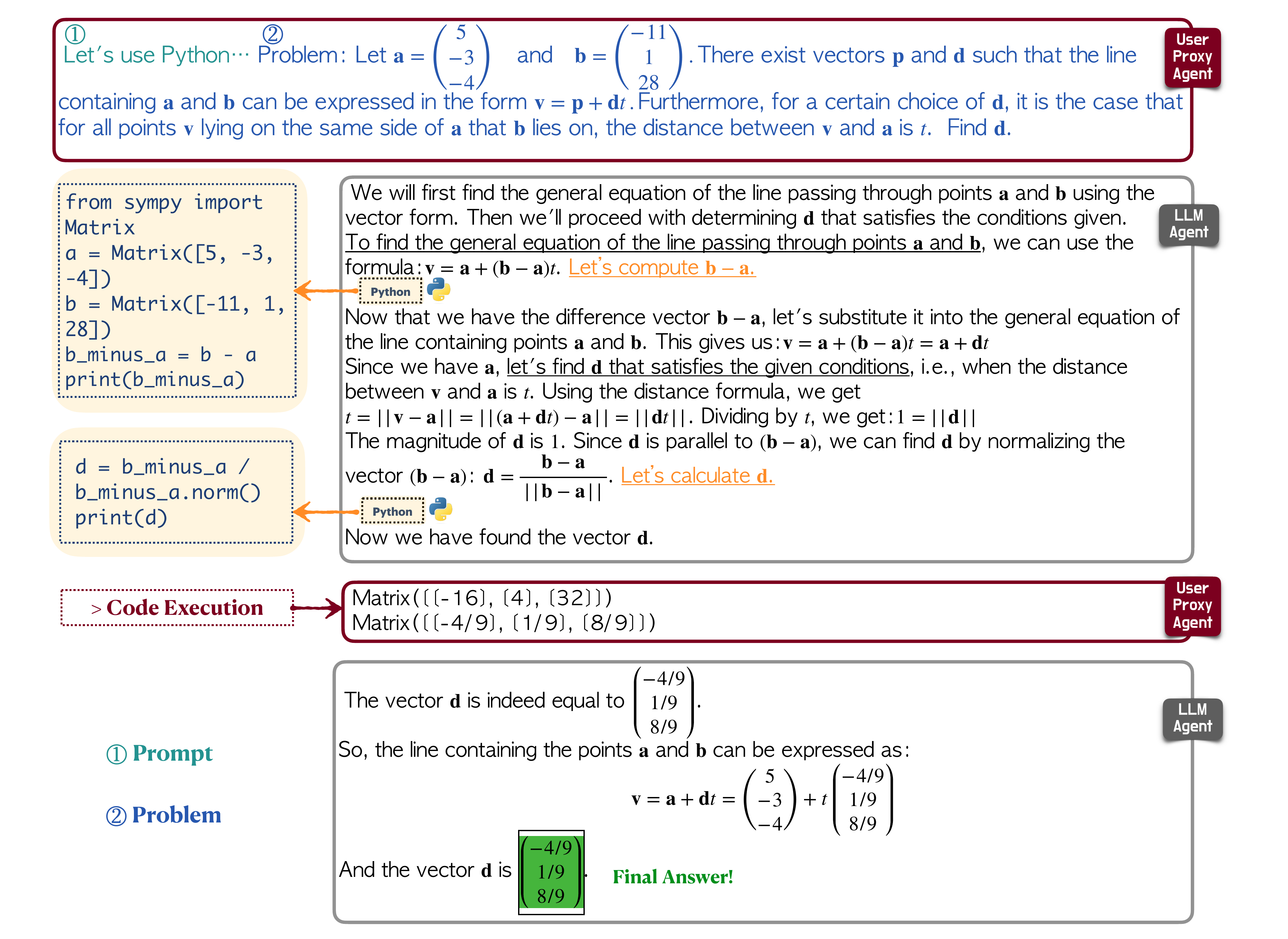}
\caption{Example of a math problem-solving process with \MathChat. The user proxy agent initiates a conversation by sending the math problem to be solved an LLM agent with preset prompt). From GPT-4's response, the user proxy agent extracts all code and executes them sequentially. Valid code from previous runs is recorded and will be executed together with the new code to reflect the step-by-step reasoning progress of the model. The results will be returned to GPT-4 and GPT-4 will continue its problem-solving process. While GPT-4 solves this problem with only one turn of user message in this example, our framework allows multi-turn conversations and additional query handling, shown in Figure~\ref{fig:queryexample}. The user proxy agent will do pattern-matching (in our case, the appearance of \texttt{\textbackslash boxed\{\}} containing a final answer) in the LLM agent's response to determine whether to end the conversation. 
}
\label{fig:example}
\vspace{-5pt}
\end{figure}

\begin{figure}[tb]
\centering
\includegraphics[width = 1\hsize]{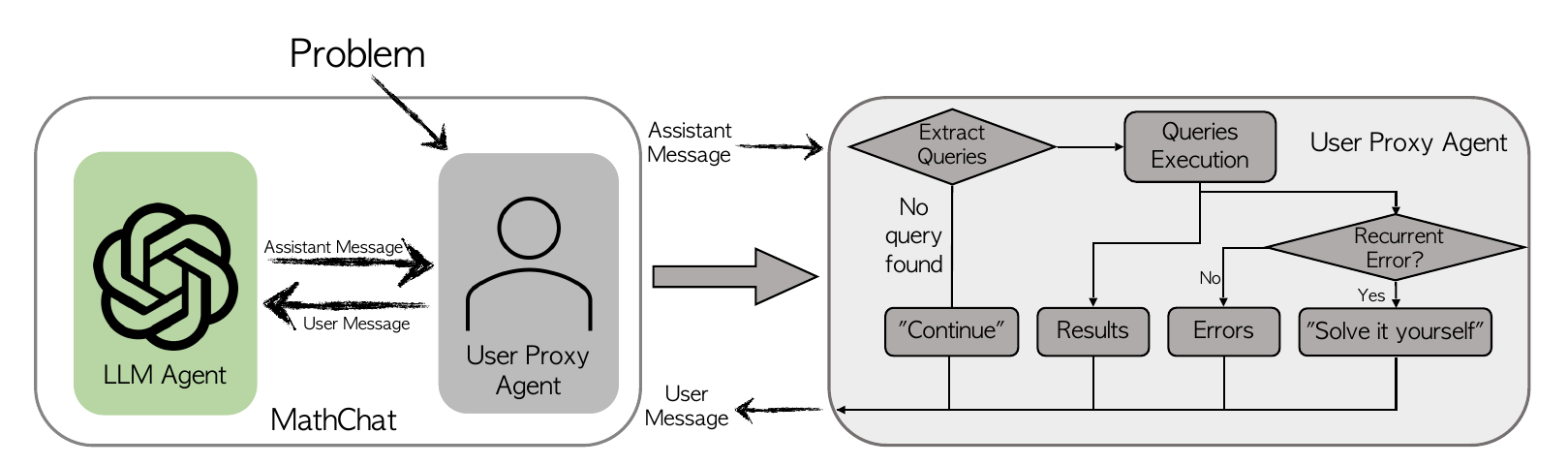}
\caption{\MathChat workflow: After a math problem is fed into \MathChat, the user proxy agent will initiate a conversation with the LLM agent to solve the problem. In each turn of interaction, the user proxy agent processes the message from the LLM agent (Assistant Message) and responds with a User Message. This process continues until the user proxy agent detects a certain pattern to end the conversation. The process in the rectangular on the right-hand side of the figure shows the inner workflow of the user proxy agent once an Assistant Message is received. It shows the functionality of executing any tool-using queries, such as Python code. It is also responsible for giving different instructions corresponding to different types of messages from the LLM agent (More in Appendix \ref{app:queryhandle}). To illustrate this feature of the proxy agent, we give a concrete example in Figure~\ref{fig:queryexample}. 
}
\vspace{-10pt}
\label{fig:mathchat_workflow}

\end{figure}

In this work, we investigate the potential of solving challenging math problems through conversations between agents. Due to the complex nature of these problems, we usually need to decompose them into multiple steps and can only make meaningful progress when all previous steps are correct. We believe conversations (together with code execution) are an ideal format, which enables iterative refining and debugging of each step. We propose \MathChat, a conversational framework tailored to chat-based LLMs, where the math problem is solved with a mock conversation between an LLM-based agent and a user proxy agent (See Figure~\ref{fig:example} for an example and Figure~\ref{fig:mathchat_workflow} for the workflow). 
We also study and incorporate effective prompting methods to instruct the LLM-based agent to solve challenging math problems more effectively.


We evaluate \MathChat with GPT-4 on the MATH dataset~\citep{hendrycks2021measuring}, a comprehensive collection of mathematical problems derived from various competitions and educational levels. We target the level-5 difficulty problems within this dataset, which primarily consist of challenging high school competition problems that even college students find difficult.  
Recognizing code execution as a major boost in performance, we compare two methods that both use Python: Program of Thoughts (PoT) prompt~\citep{chen2022program}, and Program Synthesis prompt~\citep{drori2022neural}. We also include a vanilla prompt for reference. The evaluation shows that \MathChat can further improve previous tool-using prompting methods by 6\%, and it can reach 60\% accuracy on half of the categories while having competitive performance across all categories. We also demonstrate the extensibility of \MathChat with different prompts and different tools from our experiment.
We conduct a detailed analysis of the failure reasons of all the methods evaluated. 

\vspace{-1em}
\section{Related Work} 
\vspace{-0.5em}
\label{sec:targetmath}

\paragraph{LLM Agent Systems} In the domain of LLM Agent Systems, various implementations have demonstrated the utility and diversity of multi-agent AI models. BabyAGI~\citep{babyagi} exemplifies an AI-powered task management system using multiple LLM-based agents with a static agent conversation pattern, while CAMEL~\citep{li2023camel} showcases a communicative agent framework emphasizing role-playing and autonomous cooperation. Further, research on Multi-Agent Debate~\citep{liang-arxiv2023, du2023improving} highlights the efficacy of agent debates in enhancing divergent thinking and factuality in LLMs. MetaGPT~\citep{hong2023metagpt}, a specialized application, demonstrates the use of GPTs in collaborative software development. AutoGen~\citep{wu2023autogen} is an open-source framework for creating diverse LLM applications with customizable, conversable agents using LLMs, human input, and tools.

  \vspace{-1mm}
\paragraph{Prompting Methods} 
 Creative ways of using LLMs to solve math problems have emerged lately~\citep{wang2022self, zhou2023solving, zheng2023progressive, chen2021evaluating, weng2022selfveri, wu2024stateflow}. One particular endeavor is using LLMs to offload arithmetic calculations and other basic operations involved in math problem-solving to programs~\citep{drori2022neural, chen2022program, gao2022pal}. Cumulative Reasoning~\citep{zhang2023cumulative, song2024adaptive} decomposes tasks into smaller components, streamlines the solving process, and generates thoughts in a cumulative manner. Plan \& Solve prompting \citep{wang2023plan} ask the LLM to first generate a plan and then solve it accordingly. Other general methods used to improve reasoning can also be applied to math problems: \textbf{(1)} Chain-of-thought (CoT) prompting~\citep{wei2022chain,kojima2022large} elicits step-by-step reasoning process from LLMs. \textbf{(2)} Another effective way is to prompt LLMs to solve problems in a multi-stage manner~\citep{dua2022successive, press2022measuring, creswell2022SI, long2023large, paranjape2023art, yao2022react, yang2022seqzero, long2023large, besta2023graph}. Least-to-most prompting ~\citep{zhou2022least} and Decomposed prompting~\citep{khot2022decomposed} break down a complex problem into smaller subproblems, and the subproblems are solved sequentially to reach the final solution. \textbf{(3)} Utilizing tools can significantly boost the performance of LLMs~\citep{shen2024hugginggpt, parisi2022talm}.
 ReAct~\citep{yao2022react} and ART~\citep{paranjape2023art} both use few-shot prompting to interleave step-by-step reasoning and tool-using. \textbf{(4)} Self-consistency~\citep{wang2022self}, built on top of CoT, samples several different reasoning paths for a problem and selects the answer with the majority vote.~\cite{li2022advance} extends self-consistency by training a verifier to verify the correctness of each step. By decomposing a problem-solving process, Tree-of-Thoughts~\citep{yao2023tree} proposes a set of thoughts for each intermediate step, exploring and maintaining the most promising thoughts for sequential actions.

\vspace{-1em}
\section{\MathChat: A conversational framework for math problem solving}
\vspace{-1em}
In this section, we introduce \MathChat, a conversational framework for math problem-solving.

\begin{figure}[tb]
\centering
\includegraphics[width = 1.0\hsize]{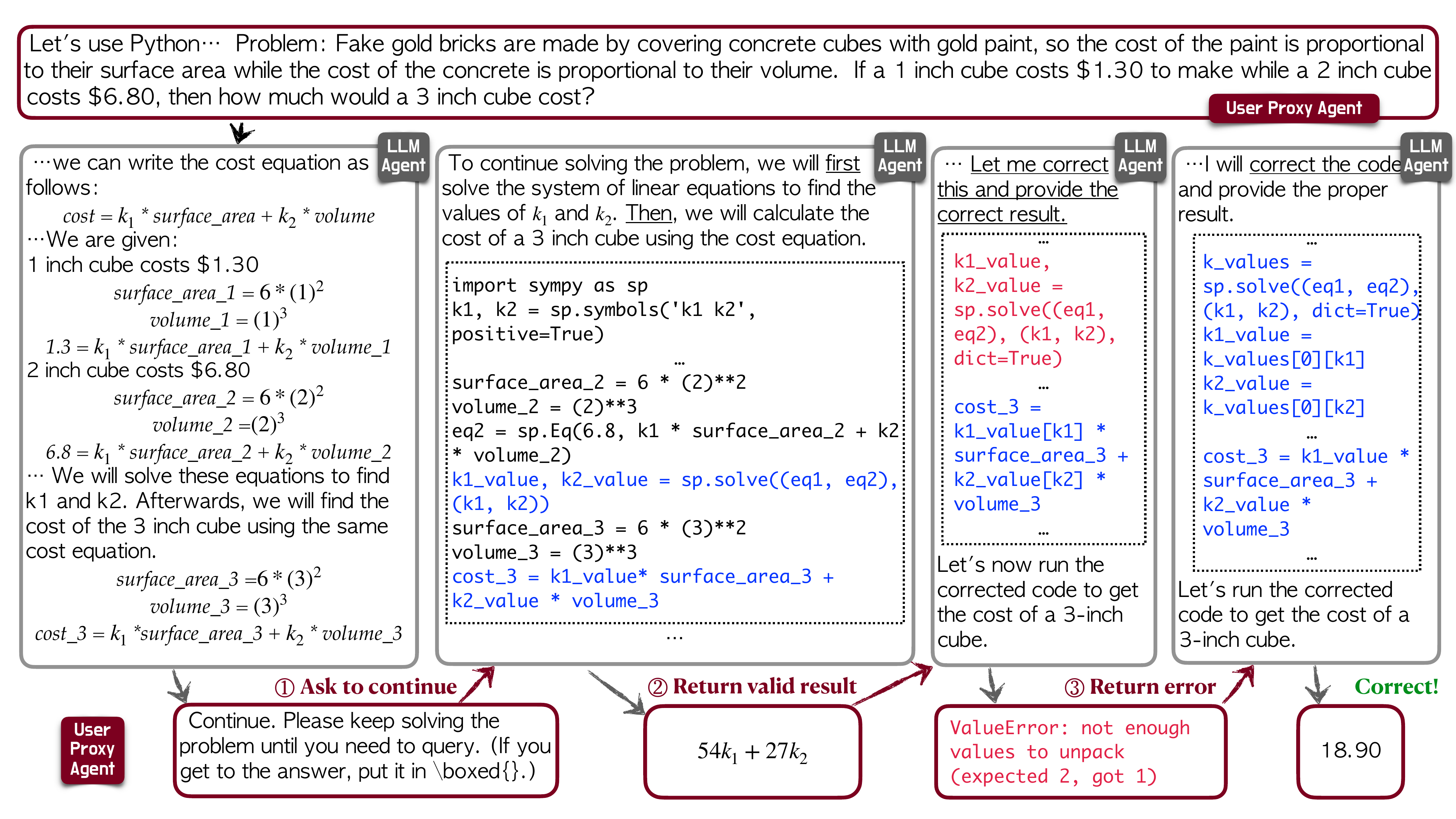}
\centering
\vspace{-5pt}
\caption{An example demonstrating how the user proxy agent handles different types of messages received from  GPT-4 in \MathChat. Specifically, the user proxy agent may respond in the following ways: (1) asking the LLM agent to continue because no code block (i.e., query) is detected; (2) returning the valid results from code execution; and (3) returning the error message from Python execution. Note that GPT-4 may change the query if the old code is undesired based on the messaged from the user proxy agent. In the last step, GPT-4 corrects the query, and the final result is returned.  
}
\label{fig:queryexample}
\end{figure}


\textbf{A conversational framework with user proxy agent.} \MathChat is a framework that simulates a mock conversation between an LLM agent (GPT-4 in our case) and a user proxy agent.  Here a user proxy agent is an agent playing the user's role in conversations with the LLM agent. In \MathChat, the LLM agent and the user proxy agent work together to solve the math problem. The workflow of this framework is presented in Figure~\ref{fig:mathchat_workflow}. The user proxy agent takes a math problem to be solved as input and would initiate a conversation with the LLM agent. The initial message from the user proxy agent consists of an initial prompt and the problem to be solved. The initial prompt is used to instruct the LLM agent to solve the problem collaboratively with the user (effectively the user proxy agent in the \MathChat system) in a certain desired manner. This framework is designed in this conversational manner in order to leverage the chat-optimized feature of state-of-the-art LLMs, e.g., GPT-4. Another distinct benefit of this framework is that it enables multi-turn dialogues, which can be particularly useful in addressing complex issues that require multi-step reasoning and tool using.

 \begin{figure}[tb]
\centering
\includegraphics[width = 1.0\hsize]{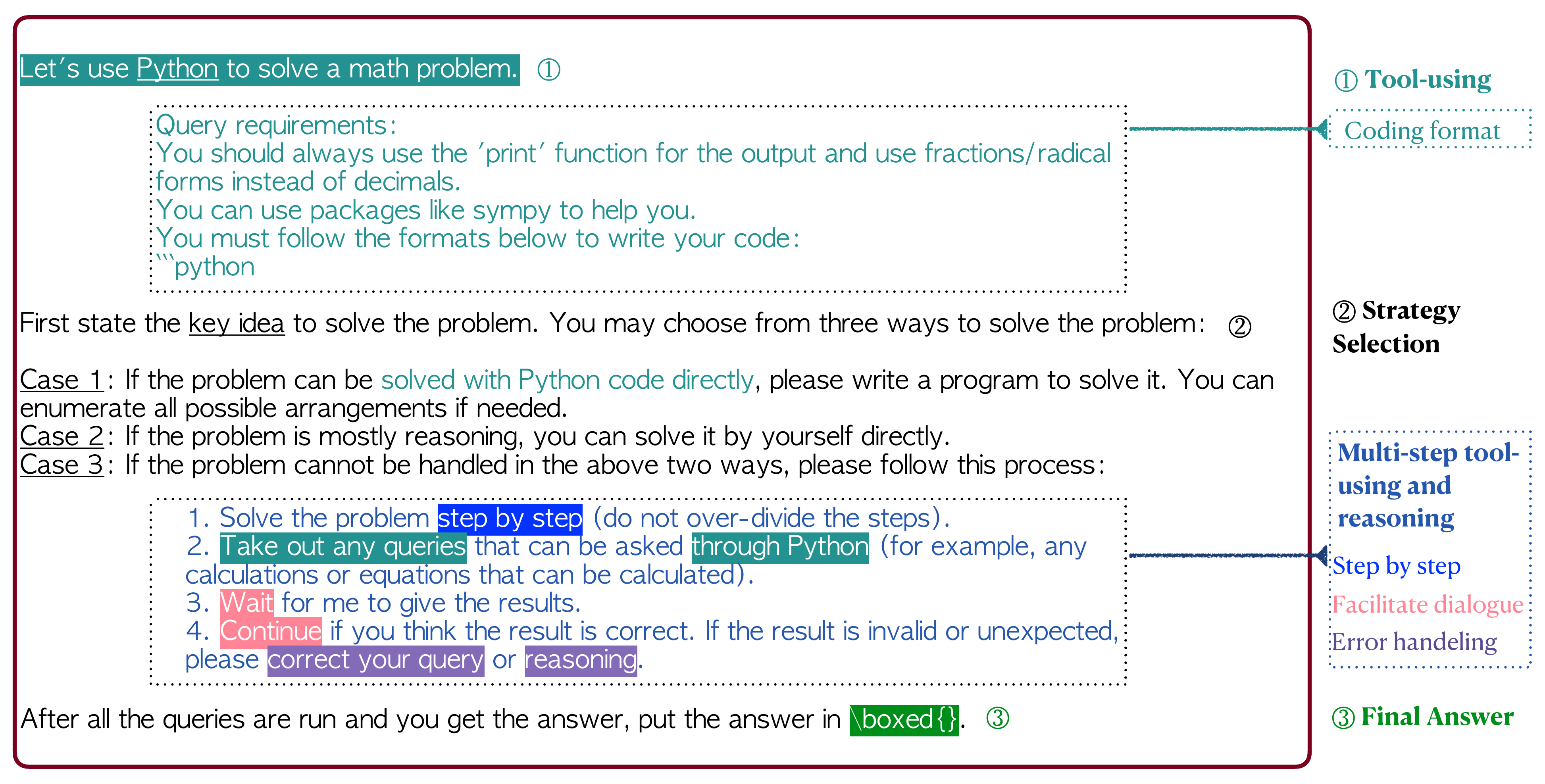}
\caption{The prompt used in the initial message of the user proxy agent in \MathChat. It instructs the LLM agent to solve a problem collaboratively with the user proxy agent in a certain way.}
\vspace{-0.5em}
\label{fig:prompt}
\end{figure}
\textbf{Prompting and tool-using in \MathChat.} With proper modifications, effective prompting methods from existing research, such as CoT and tool-using, can be integrated into the \MathChat framework. Specifically, for the prompt in the initial message, we aggregate multiple effective prompting techniques to instruct the LLM agent. We present the designed prompt in Figure~\ref{fig:prompt}, which consists of three main components.

\begin{itemize}[leftmargin=*]
    \vspace{-2mm}
    \setlength\itemsep{-0.2em}
 \item Tool-using Prompt: This component prompts the LLM to use Python programming in the correct format to tackle the problem. We use the `query requirement' subsection to specify the coding format so that the user proxy agent can parse the code and return the corresponding results. 
 \item Problem-Solving Strategy Selection Prompt: This component instructs the LLM agent to select from three possible problem-solving strategies and to perform multi-stage reasoning and tool-using in the last strategy. The problem-solving strategies include the following three cases, which cover the most effective strategies from existing literature on math problem-solving. \emph{(Case 1) Write a Python program to solve the problem directly.} This corresponds to single-stage tool-using methods similar to~\cite{gao2022pal, drori2022neural, chen2022program}. \emph{(Case 2) Solve the problem directly without Python.} 
 This strategy allows GPT-4 to exercise its inherent reasoning capacity to solve the problem at hand.
 \emph{(Case 3) Solve the problem step by step and use Python to help with math operations.} If the first two ways are not suitable, we ask the model to choose this way to solve the problem. We craft a zero-shot version of the multi-step tool-using prompt that allows the model to flexibly interleave between multi-step reasoning and Python code, similar to~\cite{yao2022react, paranjape2023art, schick2023toolformer}. In this case, we also ask the model to handle errors and unexpected results from the run of programs~\cite{ni2023lever}. 
\item Final Answer Encapsulation Prompt: This component of the prompt instructs the LLM agent to enclose the final answer in \texttt{\textbackslash boxed\{\}}, which will be used as an indicator to end the conversation. This interaction between the LLM agent and the user proxy agent will not be ended until \texttt{\textbackslash boxed\{\}} is detected or max rounds of conversations are reached.
 \end{itemize}
We acknowledge that there could be alternative ways to design the prompt. Fortunately, it is fairly easy to refine the prompt, for example, further enabling the usage of Wolfram Alpha in addition to Python, in our framework. We perform an empirical evaluation accordingly to test two alternative versions of the prompt in Section~\ref{sec:addresults}. 

\vspace{-0.8em}
\section{Evaluation}
\vspace{-0.6em}
\textbf{Dataset.} We perform evaluations on all the level-5 (the highest difficulty level) problems from the test set of MATH dataset~\cite{hendrycks2021measuring}. 
Compared to other datasets for mathematical problems such as GSM8k~\cite{cobbe2021gsm8k}, the level-5 problems are much more challenging and include the application of theorems and complex equation derivation. The MATH dataset has 7 categories of problems: Prealgebra, Algebra, Number Theory, Counting and Probability, Geometry, Intermediate Algebra, and Precalculus. In our evaluation, we remove Geometry from the evaluation to make it consistent with previous work~\cite{drori2022neural} (additional explanation in Appendix~\ref{appendix:prompt}).

\textbf{Evaluated Methods.} Most previous work uses few-shot examples to elicit the reasoning of LLMs and tool-using. It is important to select similar examples to the unanswered problem, and then annotate the examples to cover all the cases that the LLMs might encounter. A considerable amount of effort and careful consideration are required in this process. For example, \cite{khot2022decomposed, zhou2022least} relies on elaborate examples to showcase the patterns,~\cite{paranjape2023art} maintains an example library to choose examples. Note that these methods use elementary math problems and it requires even more effort to prepare and choose the examples needed for challenging math problems. 
On the other hand, multiple existing studies~\cite{openai2023gpt4,bubeck2023sparks} reveal GPT-4's remarkable capacity to follow instructions. Thus, we are interested in zero-shot prompting techniques that could enhance math-solving of GPT-4, without any example selection and annotations. Following this criterion, we evaluate our \MathChat framework with the introduced prompt and the following methods which are all zero-shot methods: vanilla prompt, Program of Thoughts~\cite{chen2022program}, and the Program Synthesis prompt from~\cite{drori2022neural}.

\begin{enumerate}[leftmargin=*]
    \vspace{-2mm}
    \setlength\itemsep{-0.2em}
 \item \textbf{Vanilla prompting:} 
 GPT-4 can perform CoT reasoning without few-shot examples. To evaluate GPT-4's performance on solving the problem directly, we use a default prompt adapted from the few-shot prompt in MATH dataset: \texttt{"Solve the problem carefully. Put the final answer in \textbackslash boxed\{\}.  \{Problem\}"}. 
 \item \textbf{Program of Thoughts (PoT):} We use the zero-shot PoT prompting from~\cite{chen2022program}, which asks a model to write a \texttt{Solver} function to solve a problem and return the final answer directly.
 \item \textbf{Program Synthesis (PS) prompting:} Similar to PoT, the Program Synthesis (PS) prompting method~\cite{drori2022neural} uses a prompt to ask the model to write a program to solve a problem: \texttt{"Write a program that answers the following question: \{Problem\}"}
\end{enumerate}

\textbf{Evaluation Details.} Hyperparameters plays critical role in determining model performance~\cite{zhang2023targeted,zhang2023hypertime}. To ensure a fair comparison, we use the default configurations from the OpenAI API on GPT-4 for all methods. In \MathChat, we allow a max round of 15 messages between GPT-4 and the user proxy agent. The agent will explicitly ask GPT-4 to solve each step by itself if it detects errors from 3 consecutive executions. To avoid extremely long responses from the user proxy agent, the agent will replace any result that exceeds 600 tokens with a warning text in the user message to ask the GPT-4 to revise the previous code.
We manually go through the answer of all the methods to count all the correct answers. For vanilla prompt, Program Synthesis, and \MathChat, we ask GPT-4 to enclose the final answer in \texttt{\textbackslash boxed\{\}}, so only the answers in the box will be extracted. For PoT, we follow the original paper to take the return of the \texttt{Solver} function as the final answer. 
\vspace{-0.5em}
\section{Results}
\vspace{-0.3em}
\subsection{Main Results}
\vspace{-0.5em}
We perform an evaluation on six categories of level-5 problems from the MATH dataset. We report the problem-solving accuracy of different methods in each category in Table~\ref{tab:main}. Compared to vanilla prompting, which shows the native capability of GPT-4, using Python with PoT or PS improves the overall accuracy by around 10\%. We can see this improvement mostly in the categories that involve more number manipulations (Counting \& Probability and Number Theory) and more challenging categories (Intermediate Algebra and Precalculus). For Algebra and Prealgebra, however, PoT and PS have little improvement or even lead to lower accuracy. Compared with PoT and PS, \MathChat can further improve the total accuracy by around 6\%, and have competitive performance across all the categories. It is worth highlighting that \MathChat improves the accuracy in the Algebra category over other methods by around 15\%. Considering all the methods, Intermediate Algebra and Precalculus can only be solved with a low accuracy rate of around 20\%. More than half of the problems from the other categories can be solved correctly by \MathChat.

\begin{table}[htb]
\centering
\begin{tabular}{@{}c|ccccccc@{}}
\toprule 
& Algebra & C.Prob & I.Alg & N.Theory & Prealg & Precalc & Total\\
\footnotesize{Problem Count} & 
\footnotesize{307} & 
\footnotesize{123} & 
\footnotesize{280} & 
\footnotesize{154} & 
\footnotesize{193} & 
\footnotesize{135} & 
\footnotesize{1192} \\ 
\midrule
\textbf{\MathChat}     &\textbf{59.93\%} & \textbf{52.03\%} & 17.85\%  & 60.39\% & \textbf{60.10\%} & \textbf{19.26}\%  & \textbf{44.71\%} \\
PoT              & 42.67\% & 50.41\% & 17.50\%  & 54.55\% & 52.33\% & 16.30\% & 37.67\% \\
PS              & 43.32\% & 44.71\% & \textbf{20.36\%}  & \textbf{61.03}\% & 55.96\% & 18.52\% & 39.60\% \\
Vanilla        & 46.58\% &25.20\%  & 2.86\% &28.57\%  &54.92\%  & 7.41\% &  28.69\%\\ 
\bottomrule
\end{tabular}

\caption{Accuracy on all the problems with difficulty level-5 from different categories of the MATH dataset with different methods.}
\label{tab:main}
\end{table}

\subsection{Additional evaluation on \MathChat with alternative prompts} 
\label{sec:addresults}
\begin{table}[htb!]
\centering
\begin{tabular}{@{}c|ccccccc@{}}
\toprule 
& Algebra & C.Prob & I.Alg & N.Theory & Prealg & Precalc & Total\\
\footnotesize{Problem Count} & 
\footnotesize{50} & 
\footnotesize{50} & 
\footnotesize{50} & 
\footnotesize{50} & 
\footnotesize{50} & 
\footnotesize{50} & 
\footnotesize{300} \\ 
\midrule
\textbf{\MathChat} w/ Two-tools  & \textbf{33} & 22 & 6  & 27 & 29 & 10 & 127\\
\textbf{\MathChat} w/ Python  & 26 & 19 & 7  & 22 & \textbf{31} & \textbf{13} & 118\\
\textbf{\MathChat}  &30 & \textbf{24} & 8  & \textbf{34} & 28 & 10  & \textbf{134} \\
PoT              & 20 & 19 & 9  & 24 & 24 & 7 & 103\\
PS              & 17 & 19 & \textbf{12}  & 31 & 26 & 5 & 110 \\
Vanilla        & 26 & 13 & 1  & 17  & 21 &  1 & 79\\ 
\bottomrule
\end{tabular}
\caption{Additional evaluation of \MathChat with two alternative prompts. 50 problems are sampled from each problem category for this evaluation. \MathChat w/Two-tools and \MathChat w/ Python are two alternative prompts.}
\label{tab:additional}
\end{table}
\MathChat allows easy incorporation of  different prompts and tools. We perform an additional evaluation to test two alternative initial prompts with \MathChat to demonstrate its extensibility. 
 (1) A simplified prompt with Python: In this alternative, we only keep the `query requirements' subsection for python coding format and the \textbf{step-by-step tool-using} (i.e., case 3) from the default prompt. 
 (2) A simplified prompt with Python and Wolfram Alpha: In this alternative, on top of alternative (1), we add Wolfram Alpha, a computational engine, as an additional tool for the LLM agent to choose from. Details of these two alternative prompts are in Appendix~\ref{appendix:prompt}. We perform an evaluation  on randomly sampled 50 examples from each of the six problem categories. We also include results from other methods on the sample problems for comparison in Table~\ref{tab:additional}. \MathChat still performs better than other methods with the two newly crafted prompts. With \MathChat, the step-by-step prompt that allows both Python and Wolfram performs the best on Algebra, while the new prompt with only Python solves the most problems on Prealgebra and Precalculus, but has a worse performance on Number Theories. Overall, \MathChat with the default prompt still performs the best.

\vspace{-0.8em}
\section{Failure Analysis}
\label{sec:failanalysis}

\begin{figure}[tb]
\centering
\includegraphics[width = 0.8\hsize]{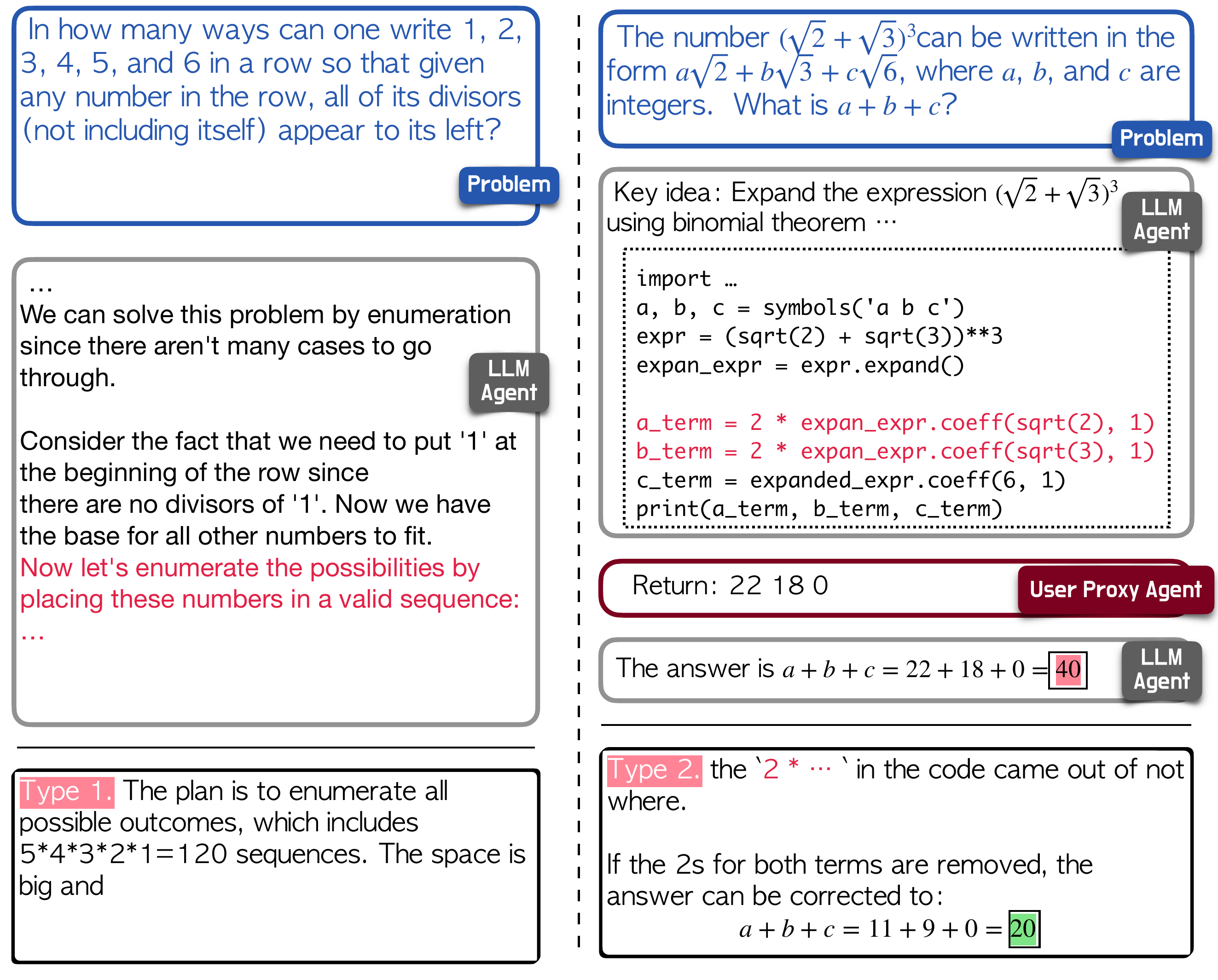}
\caption{One example is selected for each of the first two failures. 
\textbf{Type 1 failure:} in the first problem, the LLM agent fails to give a plausible plan. It chooses to enumerate all sequences, and it does not use tools to help with it. \textbf{Type 2 failure:} the second problem shows that the model fails to give the correct code to solve the problem, while it follows the problem requirements and the overall direction is correct. With minor changes to the code, the final answer can be correct. 
}
\vspace{-0.5em}
\label{fig:failure3cases}

\end{figure}

\vspace{-0.3em}
\subsection{Failure reasons}
\vspace{-0.4em}
\label{sec:failuretype}
We first summarize the failure cases according to the reasons for failure, based on the systematic math problem-solving process established by George Pólya~\cite{polya2004solve}. The process consists of (1) understanding the problem; (2) devising a plan; (3) executing the plan; (4) reviewing and extending. 
We observe failures of the following three main types. We give one example each for the two types of failures in Figure~\ref{fig:failure3cases}. More example are provided in Appendix~\ref{appendix:Failure_analysis}.


\textbf{Type 1. Failure to devise or select an appropriate plan or path to solve the problem.} This type encompasses cases where GPT-4 fails to provide a proper way to approach the problem. In these instances, the answer is consistently incorrect, even though each individual step of the calculation is accurate. 
Failure cases of this type are typically tricky problems that even math experts find challenging.

\textbf{Type 2. Failure to flawlessly execute the devised plan.} Math problem-solving requires rigorous and precise reasoning steps. A minor error in calculation or symbol manipulation could be fatal and lead to a wrong answer. This type of error is considered `minor' because they are easier to be fixed. 
This type of error contributes to a fair amount of failures, where the overall direction of the problem-solving is correct, but one mistake in a basic derivation leads to the wrong answer. Note that an error for Python execution is also included in this type, where GPT-4 fails to write a runnable code leading to the error.

\textbf{Type 3. Other technical errors.} There are other technical errors causing the failure. One example of such failure is lack of information due to the removal of ASY code.

\vspace{-0.2em}
\subsection{Failure cases using different methods on GPT-4}
\vspace{-0.1em}
\label{sec:fail_diff_ways}

\begin{figure}[!tb]
\centering
\includegraphics[width = 1\hsize]{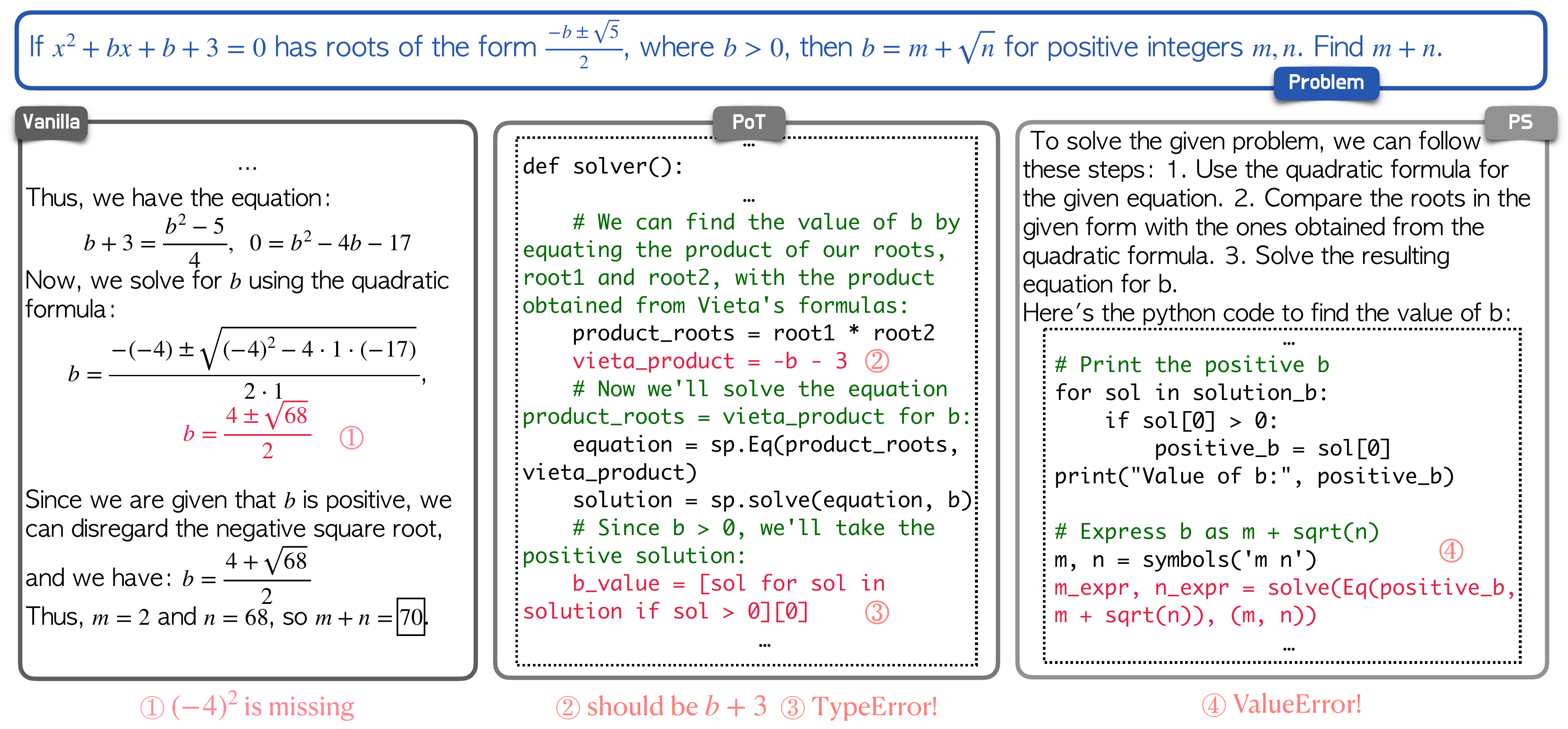}
\caption{An example where \MathChat is correct and others fail. All other methods fail due to Type 2 failure. 1. Vanilla prompt: when calculating $b$, didn't include $-4^2$. 2. PoT:
it first calculates \texttt{vieta\_product} wrong, even is this is corrected, another \texttt{TyperError} will occur. 3. PS: it solves for $b$ correctly, but gets an \texttt{ValueError} when using the program to extract $m$ and $n$.}
\vspace{-0.6em}
\label{fig:mathchat_correct}
\end{figure}


In Table~\ref{tab:method_succeed}, we present the frequency of successful outcomes for each method (represented in each row), while all other methods fail, categorized according to different problem instances. This table serves to highlight the distinct advantage that a particular method exhibits across various problem categories. Similarly, in Table~\ref{tab:method_fail}, we summarize the frequency of instances where one method fails while all other methods succeed. A high number in this table signifies the unique disadvantage of the method in question.

These statistics demonstrate the robustness of \MathChat in comparison to other methods. \MathChat leverages conversation to enhance error correction when utilizing external tools, thereby hypothesizing a reduction in failures within the third type.

\begin{table}[htb!]
\centering
\begin{tabular}{@{}c|cccccc|c@{}}
\toprule[1.5pt]
& Algebra & C.Prob & I.Alg & N.Theory & Prealg & Precalc & Total\\
\midrule
\MathChat     &\textbf{27} & 8 & 21  & \textbf{13} & 6 & \textbf{9}  & 
\textbf{84} \\
PoT          & 11 & \textbf{9}  & 19  & 6 & 3 & 5 & 53 \\
PS         & 12 & 6 & \textbf{22}  & 11 & \textbf{10} & 8 & 69 \\
Vanilla    & 12 & 4  & 5 & 3  & \textbf{10}  & 3 &  37 \\ 
\bottomrule[1.5pt]
\end{tabular}
\caption{The number of problems where one method succeeds, and all the other methods fail (the higher the better for the concerned method in each row).}
\label{tab:method_succeed}
\end{table}

\begin{table}[htb!]
\centering
\begin{tabular}{@{}c|cccccc|c@{}}
\toprule[1.5pt] 
& Algebra & C.Prob & I.Alg & N.Theory & Prealg & Precalc & Total\\
\midrule
\MathChat    &\textbf{6} & \textbf{2} & \textbf{0}  & \textbf{5} & \textbf{4} & 1  & 
\textbf{18} \\
PoT          & 22 & 5  & 0  & 6 & 18 & 2 & 53 \\
PS         & 17 & 5 & 1  & \textbf{5} & 14 & \textbf{0} & 42 \\
Vanilla    & 16 & 19  & 11 & 28  & 19  & 5 &  98 \\ 
\bottomrule[1.5pt]
\end{tabular}
\caption{The number of problems where one method fails and all the other methods succeed (the lower, the better for the concerned method in each row).}
\label{tab:method_fail}
\end{table}

We take one example from each table to analyze the failures of these methods. We first take an Algebra problem that \MathChat succeeds but others fail to analyze the failures of other methods (Figure \ref{fig:mathchat_correct}) 
For this problem, other methods fail to execute this plan without any mistakes, causing the second type of failure. While vanilla prompting has a calculation error, the other two methods get execution errors from running the code. We run these methods three more times and they still fail to solve the problem. From Table~\ref{tab:method_fail}, we take the only Precalculus instance that \MathChat is wrong while all the other methods are correct. Through investigating, we find that \MathChat gives longer solutions to the problem than all the other methods, and also contains Type 2 failures. This problem might indicate a potential correlation between the accuracy and length of responses. We present more details in investigating this possible correlation and also the Precalculus example in Appendix~\ref{appendix:Failure_analysis}.


\vspace{-0.7em}
\section{Summary and future work}
\vspace{-0.5em}
\subsection{Summary}
\vspace{-0.1em}
In this paper, we introduce \MathChat, a conversational framework to solve math problems with the collaboration of an LLM agent and a user proxy agent. \MathChat is designed for chat-optimized models like GPT-4, and it is extensible to be used with different prompts and different tools with minimal effort. Based on the framework, we also derive a prompt that aggregates previous prompting techniques to be used on \MathChat. Our evaluation of level-5 problems from the MATH dataset demonstrates the effectiveness of \MathChat to solve more complex and challenging problems. Despite its improvements over previous methods, the results show that complex math problems is still challenging for recent powerful LLMs, like GPT-4, even with help from external tools. We discuss potential directions to further improve math problem-solving below.


\vspace{-0.4em}
\subsection{Enhanced Agent Specialization in Problem Solving} 
\vspace{-0.1em}
\textit{The Society of Mind}~\cite{minsky1988society} posits that intelligence emerges from the interaction of relatively simple agents. In \MathChat, where two agents collaborate to solve math problems, the behavior of the LLM agent, guided by specific instructions for task completion and decision-making, shows significant variance. To enhance consistency and effectiveness, it could be beneficial to decompose this process into specialized tasks, each handled by a dedicated agent. One agent, for instance, could focus on comprehending and developing initial solutions, while another evaluates the most suitable problem-solving strategy for the given problem.

Other than decomposing the solving process, it is possible to categorize problems by type, difficulty level, or other aspects, and accordingly select the most effective agent (prompting method) for each category. Our analysis in Section~\ref{sec:fail_diff_ways} demonstrates that various methods exhibit distinct advantages depending on the problem type.
This approach is akin to the Mixture of Experts model~\cite{shazeer2017outrageously}, where specific prompting strategies are used in place of sub-neural networks, and it also aligns with the concept of prompting chaining \cite{wu2022promptchainer}, which involves classifying a task under various scenarios for targeted resolution.


\vspace{-0.5em}
\subsection{Assistance in Human problem-solving} 
\vspace{-0.1em}
 While LLMs is showing great potential to aid in human problem-solving, we recognize that much work remains in developing a reliable LLM-based problem-solving assistant. When conducting failure analysis in Section~\ref{sec:failanalysis}, we can spot calculation errors in LLM responses easily but may struggle with identifying logical or factual inaccuracies, especially with unfamiliar concepts. This could lead to potential misinformation, especially for students who are learning new concepts and have weaker judgements. Although our evaluation indicates that incorporating Python enhances LLM problem-solving abilities, relying solely on Python or LLMs has limitations, as Python solutions (such as brute-force or simulations) may not suit human learning needs.

 A possible mitigation would be to verify each step of the solving process with external tools and established knowledge. When the LLM generates each intermediate step to solve a problem, Python can be used to check for calculation errors in the step, and external databases can be consulted to validate any theorems mentioned.





\bibliography{iclr2024_conference}
\bibliographystyle{iclr2024_conference}

\appendix
\newpage
\section{Supplementary Details on the User Proxy Agent}
\label{app:queryhandle}
The user proxy agent in \MathChat takes a problem and put it in a message with an initial prompt, and sends the message to the LLM agent. Then the agent is responsible for extracting and executing queries and also providing additional guidance. Here are all functionalities of the user proxy agent (the workflow is shown in Figure~\ref{fig:mathchat_workflow}):
\begin{enumerate}[leftmargin=2em]
    \item \textbf{Extract Queries:} The user proxy agent needs to match the pattern specified in the initial message to extract all tool-using queries. With our designed prompt, the agent matches all code blocks in the message and extracts the code.
    
    \item \textbf{"Continue":} If no query is detected in the message, the agent will send this message to the LLM agent: \texttt{"Continue. Please keep solving the problem until you need to query. (If you get to the answer, put it in \texttt{\textbackslash boxed\{\}}."}. This asks the agent to keep solving the problem and reminds it to end the conversation by putting the answer in the box.
    
    \item \textbf{Query Execution:} Any tool-using queries extracted will be executed sequentially. For Python, we set the time limit to be 5 seconds for execution. As shown in Figure \ref{fig:example}, the previous valid code is recorded. All the execution results will be concatenated sequentially (including errors).

    \item \textbf{Recurrent Error detection:} If LLM agent sends 3 consecutive errors, the user proxy agent will replace the third error message with this message: \texttt{"Please revisit the problem statement and your reasoning. If you think this step is correct, solve it yourself and continue the next step. Otherwise, correct this step."}. To avoid sticking to this error, the LLM agent is asked to solve this step without tools and move on.
    \item \textbf{Repetitive results:} This is not shown in the workflow, but the agent also detects another situation where the LLM agent gives the same tool-using query from the last one or the result is the same as the last query. Then the message is appended to the execution result to remind the agent to avoid giving the same queries: \texttt{"Your query or result is same from the last, please try a new approach."}.
    \item  \textbf{Long query results:} It is possible that LLM agent requires a query result that is too long to be passed back (such as long results from the print function in a for loop in Python). The proxy agent will replace any query result that is longer than 2000 chars (approximately 600 tokens) with this message: \texttt{"Your requested query response is too long. You might have made a mistake. Please revise your reasoning and query."}.
\end{enumerate}

In \MathChat, if the tool-using query and the end indicator are detected in the same message, the result from the query will be returned, and the conversation will continue. This is to prevent early stops where the LLM agent predicts the execution result and puts it in a box other than waiting for the result.

\section{Supplementary Details on Experiment Settings}
\label{appendix:prompt}

Rational in removing the geometry problems from testing: Most geometry problems from this dataset contain an Asymptote code to plot the figure. But the currently available version of GPT-4 cannot accept image input. If the raw code is included, it can leak information to the model through exact numbers for the coordinates. Taking these issues into consideration, we skip the evaluation on Geometry problems and remove ASY code from all the other categories (though this could result in a lack of enough information for some problems). The correct answer to each problem is deterministic and is enclosed in \texttt{\textbackslash boxed\{\}} in the dataset as ground truth (but not disclosed to the methods solving the problem).

The code is in this \href{https://github.com/yiranwu0/MathChat}{GitHub} repository.
In our experiment, we use the default configuration from OpenAI, specifically \texttt{temperature=1}, and \texttt{max\_token=inf} (See \href{https://platform.openai.com/docs/api-reference/chat/create}{OpenAI API Reference} for more details). We use the system message "You are a helpful assistant" for vanilla prompt, PS, and \MathChat. For PoT, we do not add this system message, since our evaluation shows that PoT without system message has a better performance. We discuss the effect of system message below in Section~\ref{app:sup_result}.

\begin{figure}[!tb]

\centering
\includegraphics[width = 1\hsize]{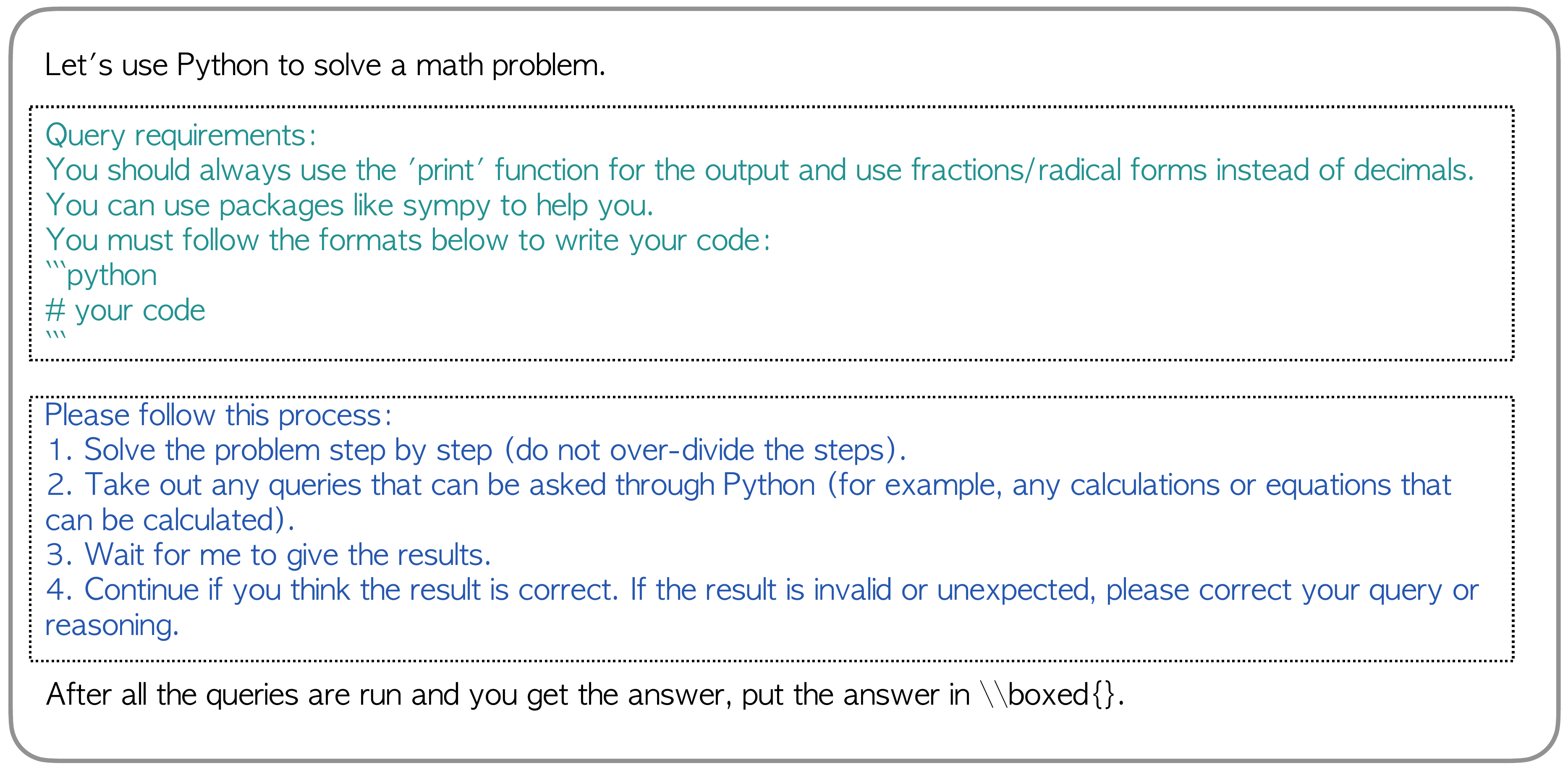}
\caption{The Python prompt used on \MathChat from Section~\ref{sec:addresults}.}
\label{fig:python_prompt}
\end{figure}

\begin{figure}[!tb]

\centering
\includegraphics[width = 1\hsize]{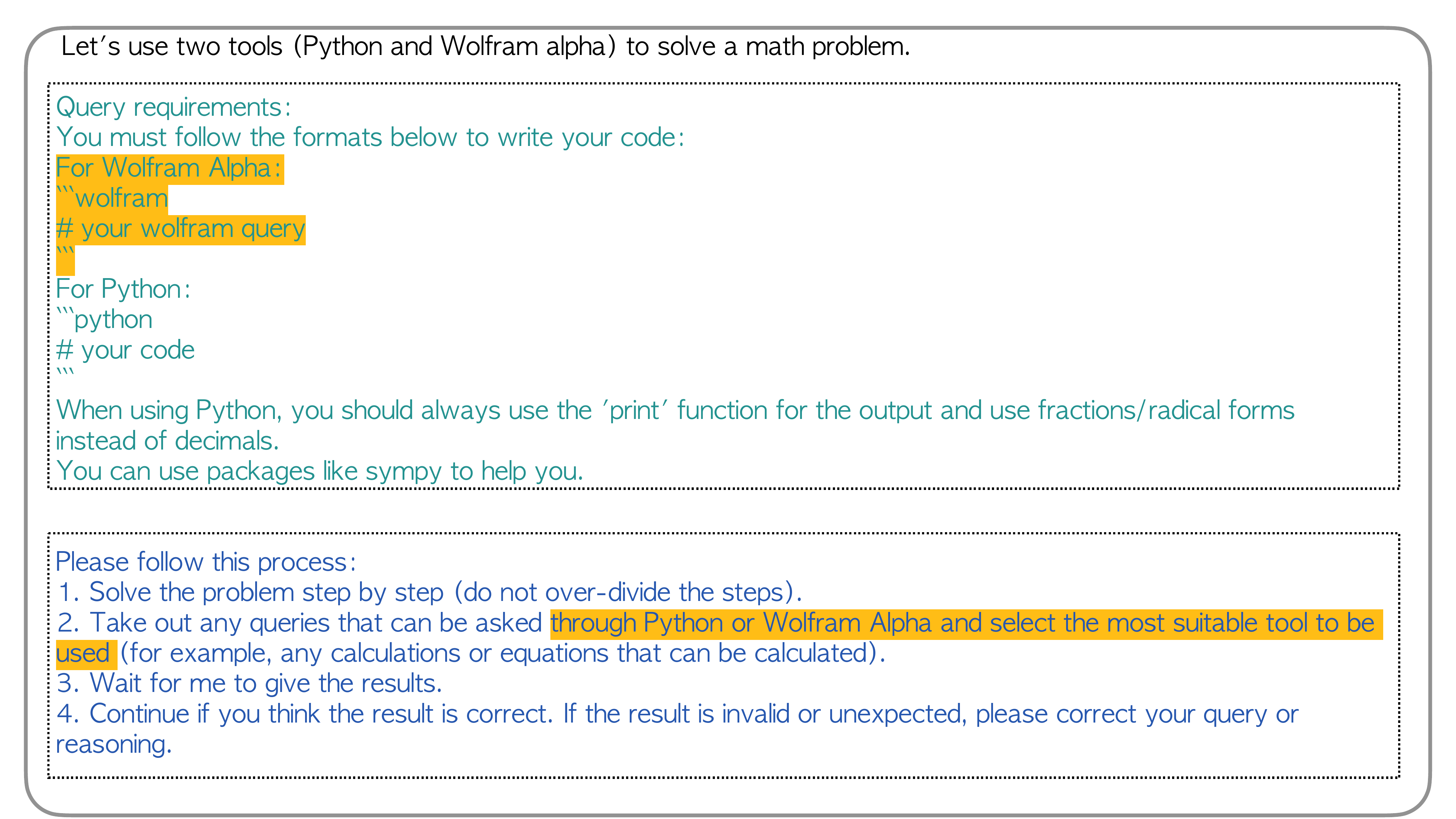}
\caption{The Two-tools prompt used on \MathChat from Section~\ref{sec:addresults}. The added requirement compared to the Python prompt is highlighted in yellow. This prompt allows the LLM agent to choose from Python or Wolfram Alpha.}
\label{fig:two_tools_prompt}
\end{figure}

\begin{figure}[!tb]

\centering
\includegraphics[width = 1\hsize]{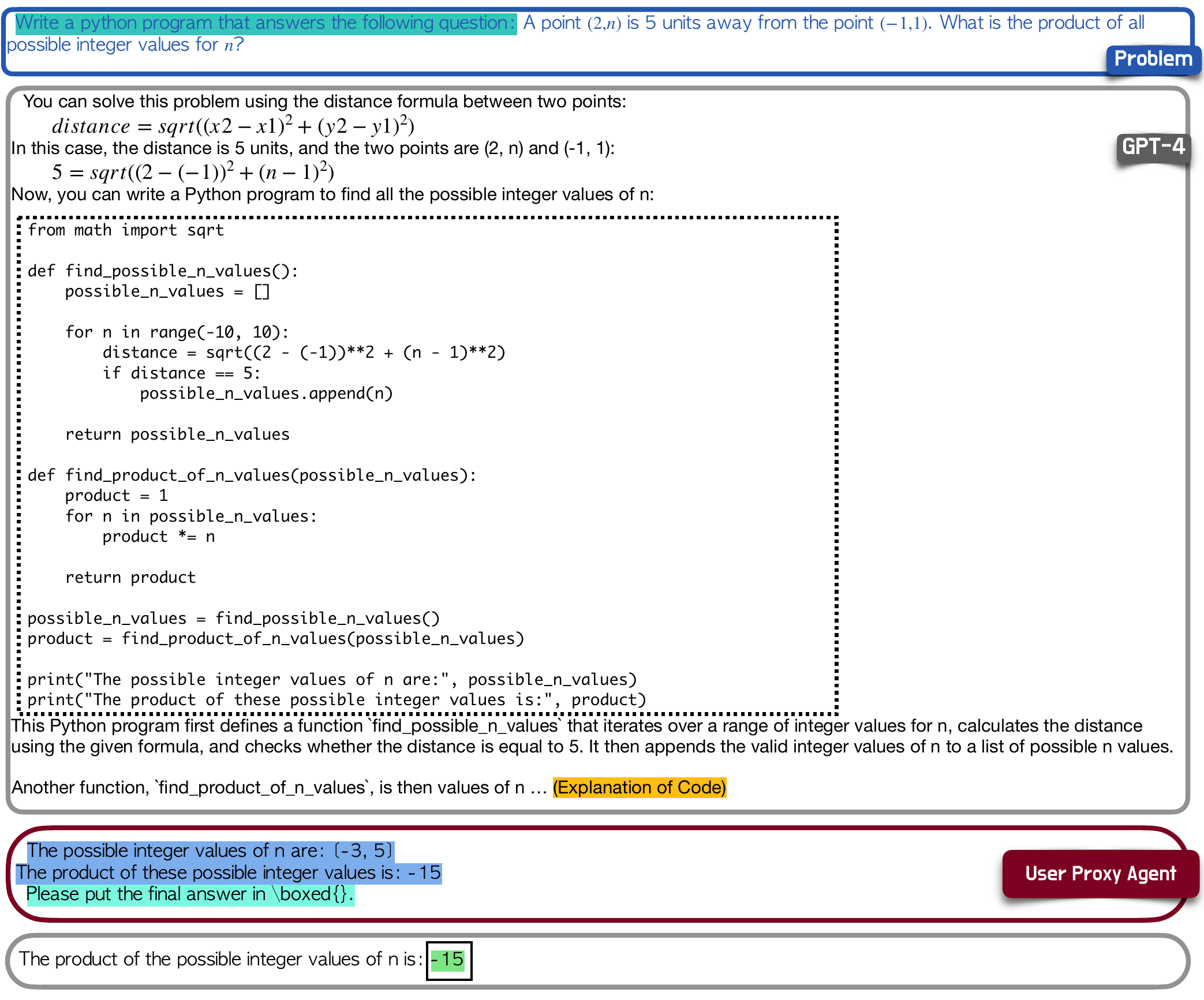}
\caption{An example of the process of PS. The query result will be returned to the LLM assistant and ask it to put the answer in box. The process of PS is exactly the same as \MathChat when the agent in \MathChat chooses to solve the problem with one Python program.}
\label{fig:example_ps}
\end{figure}

\begin{figure}[!tb]

\centering
\includegraphics[width = 0.8\hsize]{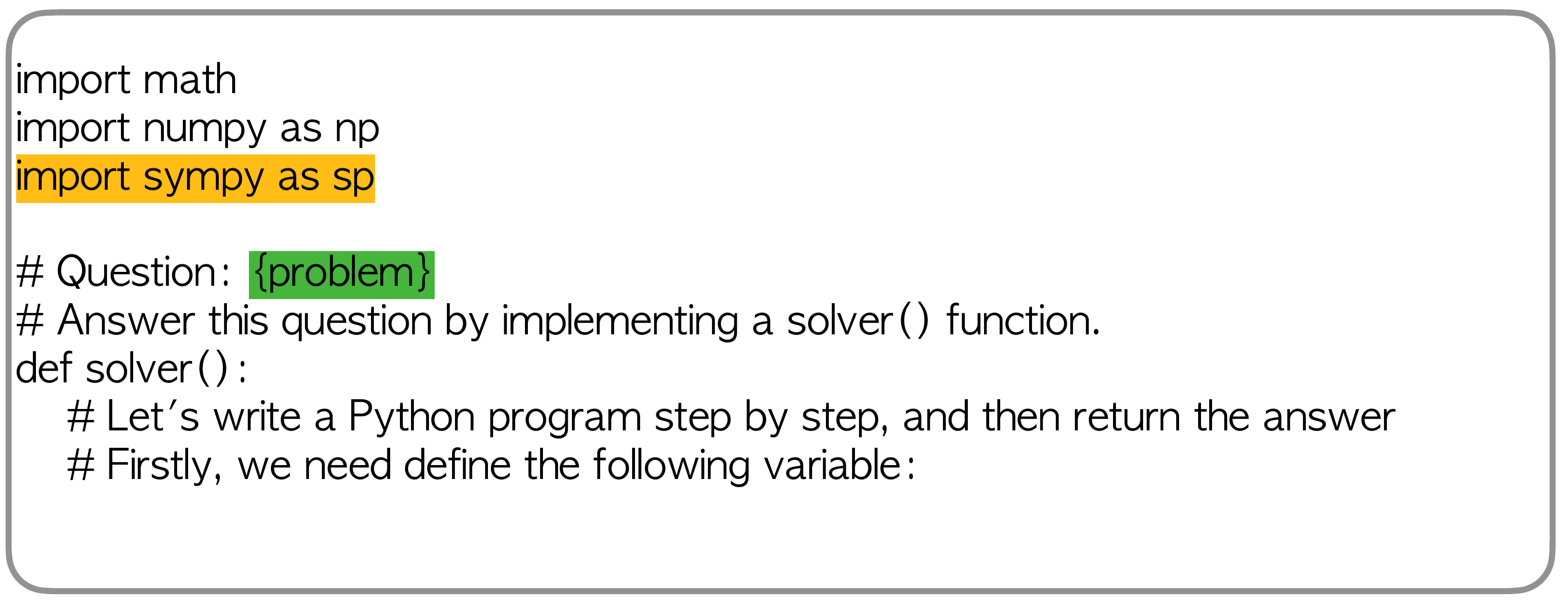}
\caption{PoT prompt. Comparing to the original prompt from \cite{chen2022program}, we add \texttt{"import sympy as sp"} that gives the LLM agent hint to use the sympy library. The placeholder \texttt{"\{problem\}"} will be replaced with the actual problem.} 
\label{fig:pot_prompt}
\end{figure}

Here is the prompts for PoT \cite{chen2022program}, PS \cite{drori2022neural}, and the additional two prompts we designed:
\begin{itemize}[leftmargin=2em]
    \item \textbf{Program of Thoughts (PoT).} See Figure~\ref{fig:pot_prompt}. The whole prompt uses the Python code format, where information such as the problem or instructions is in the comments.
    \item \textbf{Program Synthesis (PS).} The prompt for PS is \texttt{"Write a program that answers the following question: \{Problem\}"}. Since the definition of "program" is unclear and sometimes the LLM agent won't write code to solve the problem, we add the keyword 'Python' in front of 'program'. After the message is returned, we used the proxy agent to return the result (by default, GPT-4 would return the code in the code block). Then we send another message to the model with the Python execution result and ask it to enclose the final answer: \texttt{\{Return from Python\}. Please put the final answer in \texttt{\textbackslash boxed\{\}}.} See Figure \ref{fig:example_ps} for an example of the whole process.

    \item \textbf{Python prompt (w/ \MathChat).} See Figure~\ref{fig:python_prompt}.

    \item \textbf{Two-tools prompt (w/ \MathChat).} See Figure~\ref{fig:two_tools_prompt}.
\end{itemize}

\section{Supplementary Experiments and Results}
\label{app:sup_result}
We further evaluate a vanilla few-shot prompt, PoT with and without system message, and Vanilla prompt with and without system message on the randomly selected 50 problems from each category and present the results in Figure~\ref{app:addtional}. 

In the few-shot prompt, we randomly select 3 level-5 problem-solution pairs from the train set. These examples are selected from each category and are used for all the problems from that category. The vanilla few-shot prompt starts with \texttt{"Solve the problem carefully. Put the final answer in \textbackslash boxed\{\}} just like the vanilla prompt, and then three \texttt{"Problem: ... Solution: ..."} pairs are attached. Compared with the vanilla prompt, adding three additional examples does not make any obvious difference, and the overall performance is slightly worse.

From our experiment, we also notice that the system message affects the performance of the LLM agent. However, the impact significantly differs between methods. As shown in Table~\ref{app:addtional}, using a system message is crucial for the Vanilla prompt: adding the system message doubles the success rate compared to the one without a system message. However, for PoT, adding the system message only slightly increases the performance. We add a further evaluation on all the level-5 problems and find the PoT with the system message has an overall accuracy of 35.82\%, which is lower than the accuracy of PoT without the system message (37.67\% as shown in the main results in Table~\ref{tab:main}). 
We hypothesize the difference in the prompt format across different methods is the reason for this behavior. 
The method with the Vanilla prompt imitates a conversation between the LLM agent and humans via natural language, but PoT prompt is in Python code format, which explicitly directs the model for code completion. Thus, the system message "you are a helpful assistant" is more suitable for Vanilla prompt but doesn't align with PoT. More investigation is needed to understand the effect of system messages.



\begin{table}[htb!]
\centering
\begin{tabular}{@{}c|ccccccc@{}}
\toprule 
& Algebra & C.Prob & I.Alg & N.Theory & Prealg & Precalc & Total\\
\footnotesize{Problem Count} & 
\footnotesize{50} & 
\footnotesize{50} & 
\footnotesize{50} & 
\footnotesize{50} & 
\footnotesize{50} & 
\footnotesize{50} & 
\footnotesize{300} \\ 
\midrule
\MathChat  &30 & 24 & 8  & 34 & 28 & 10  & 134 \\
PS              & 17 & 19 & 12 & 31 & 26 & 5 & 110 \\
PoT w/o sys       & 20 & 19 & 9  & 24 & 24 & 7 & 103\\
PoT w/ sys       & 18 & 23 & 9  & 23 & 29 & 7 & 109\\
Vanilla w/o sys   & 14 & 4 & 0  & 4 & 13 & 1 & 35 \\
Vanilla w/ sys   & 26 & 13 & 1  & 17  & 21 &  1 & 79\\ 
Few-shot (k=3)     & 21 & 6 & 2  & 18 & 24 & 1 & 72\\
\bottomrule
\end{tabular}
\caption{Results for few-shot prompt, PoT w/ and w/o system message, Vanilla prompt w/ and w/o system message.}
\label{app:addtional}
\end{table}

\section{Supplementary Failure Analysis}
\label{appendix:Failure_analysis}

\subsection{Failure under different forms of problem-solving processes in \MathChat}
\begin{figure}[!tb]
\label{fig:correct_rate}
\centering
\includegraphics[width = 0.7\hsize]{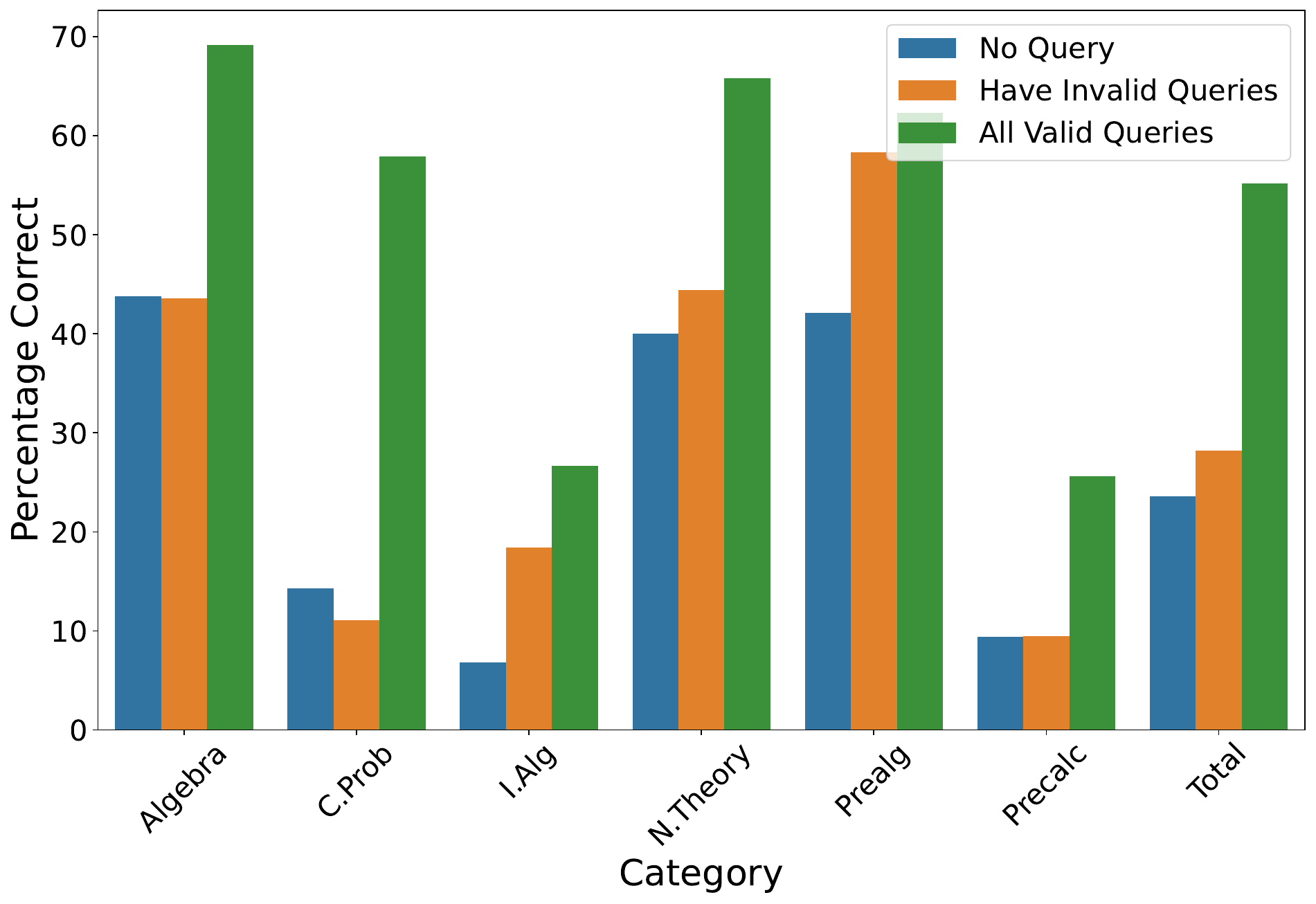}
\caption{Success rate of \MathChat under different forms of problem-solving processes: 1. the LLM agent solves the problem without making any tool-using queries. 2. The agent makes queries and has at least one invalid query 3. All queries made are valid. }
\label{app:diff_situ}
\end{figure}

The default prompt in \MathChat allows the LLM agent to choose from different forms of problem-solving processes to solve the problem, and we investigate how choosing different forms could affect the performance. We plot the correct rate when a problem from each category is solved with three forms of problem-solving approaches depending on the existence and validity of the query in the generated solution in Figure~\ref{app:diff_situ}: 1. The LLM agent doesn't make any tool-using queries (Python) when solving the problem. 2. The agent makes one or more queries, but at least one query is invalid. 3. The agent makes all valid queries. It is shown in the plot that using Python correctly could significantly increase the correct rate, while doesn't use Python is worse than using Python but having invalid queries. The results in Figure~\ref{app:diff_situ} show that especially for intermediate algebra and prealgebra, the gap in accuracy between "using no query" and "have invalid query" is large, indicating that using Python is very helpful to solve problems from the two categories. 



\begin{figure}[!tb]
\centering
\includegraphics[width = 1\hsize]{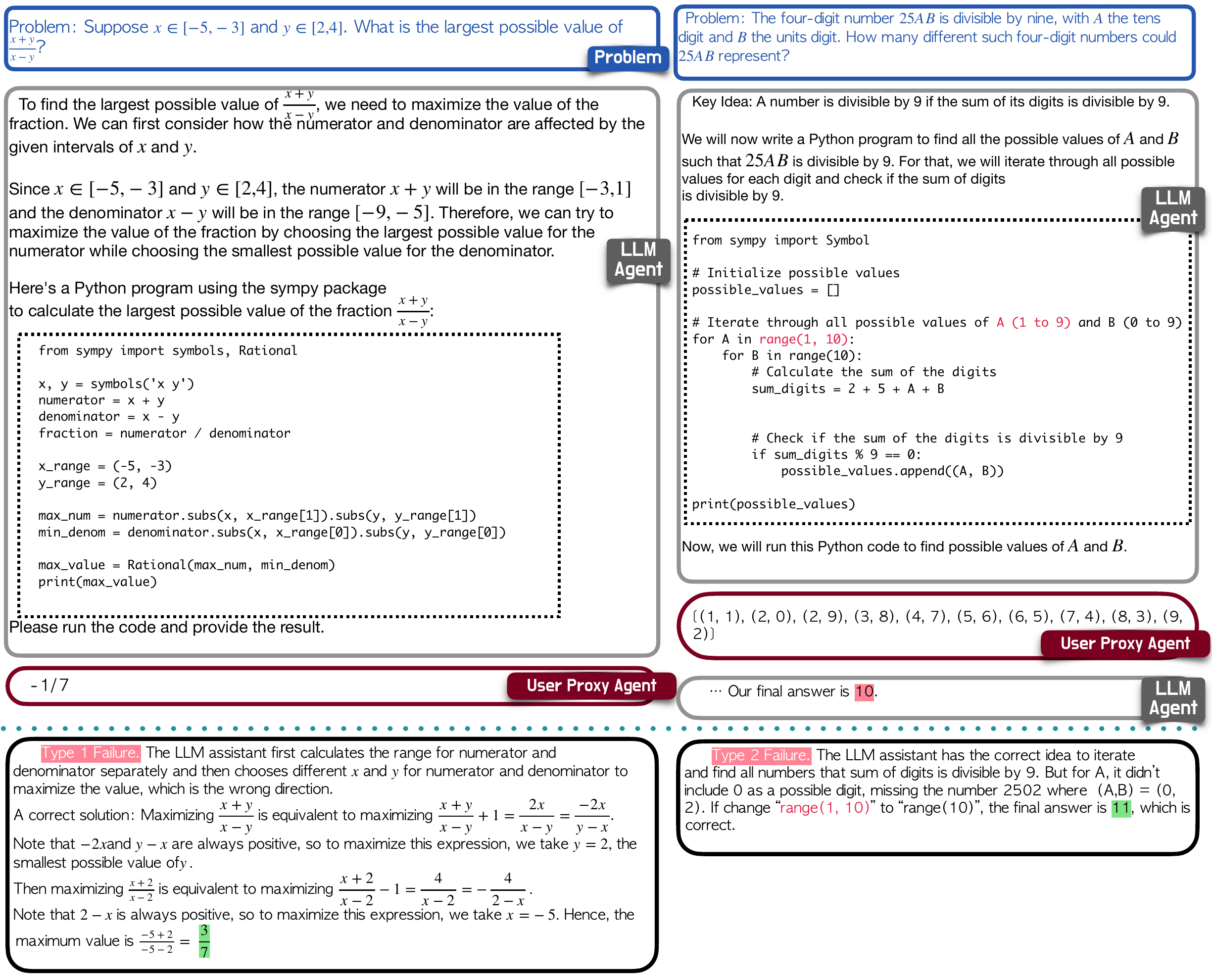}
\caption{Additional example of Type 1 failure (Fail to devise a proper plan) and Type 2 failure (Fail to execute the plan flawlessly).}
\label{app:addtype1type2}
\end{figure}


\begin{figure}[!tb]
\centering
\includegraphics[width = 1\hsize]{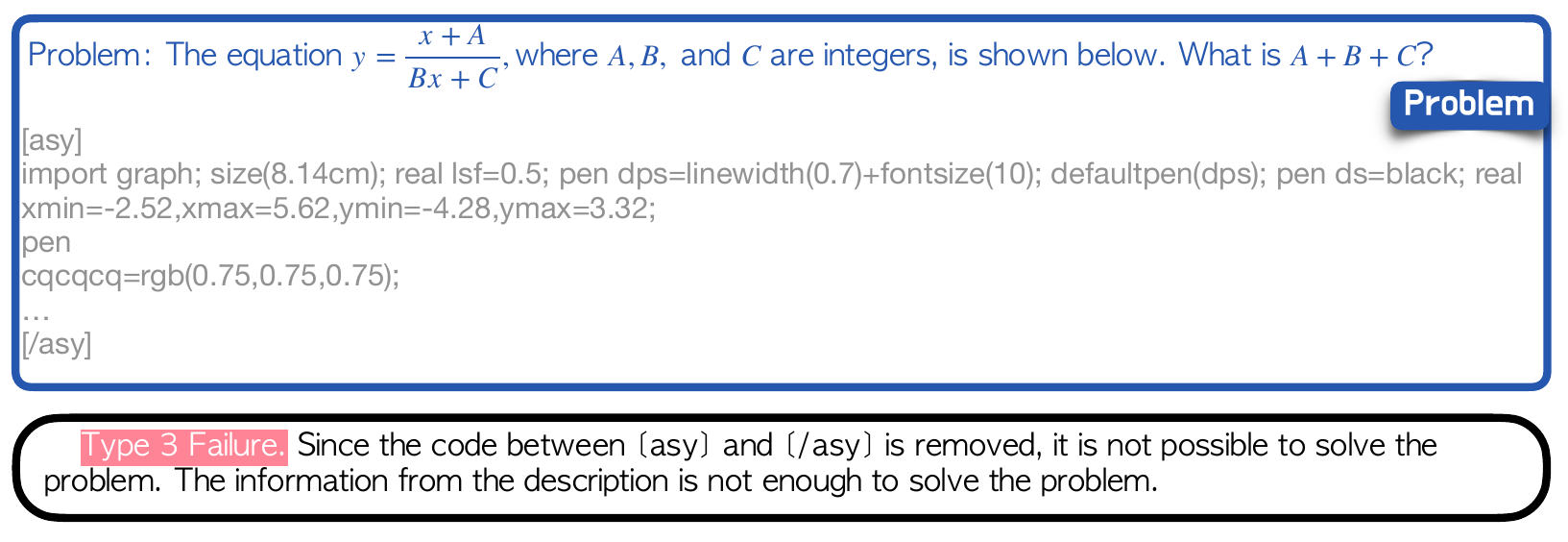}
\caption{An example of Type 3 failure where the ASY code is removed. }
\label{app:type3}
\end{figure}

\subsection{Examples of 3 types of failures}
In Section~\ref{sec:failuretype}, we summarize 3 main type of failures: type 1: failure to devise an appropriate plan. type 2: failure to flawlessly execute the plan. type 3: other technical errors. We give one additional example for type 1 error and type 2 error, and an example where the removal of ASY code leads to a leak of information (Figure~\ref{app:addtype1type2}, Figure~\ref{app:type3}). We note that among all the problems,  the ASY code from 72 problems are removed, but 12 problems can still be solved correctly. 

\subsection{Failure under different methods}
\begin{figure}[tb]
\centering
\includegraphics[width = 1\hsize]{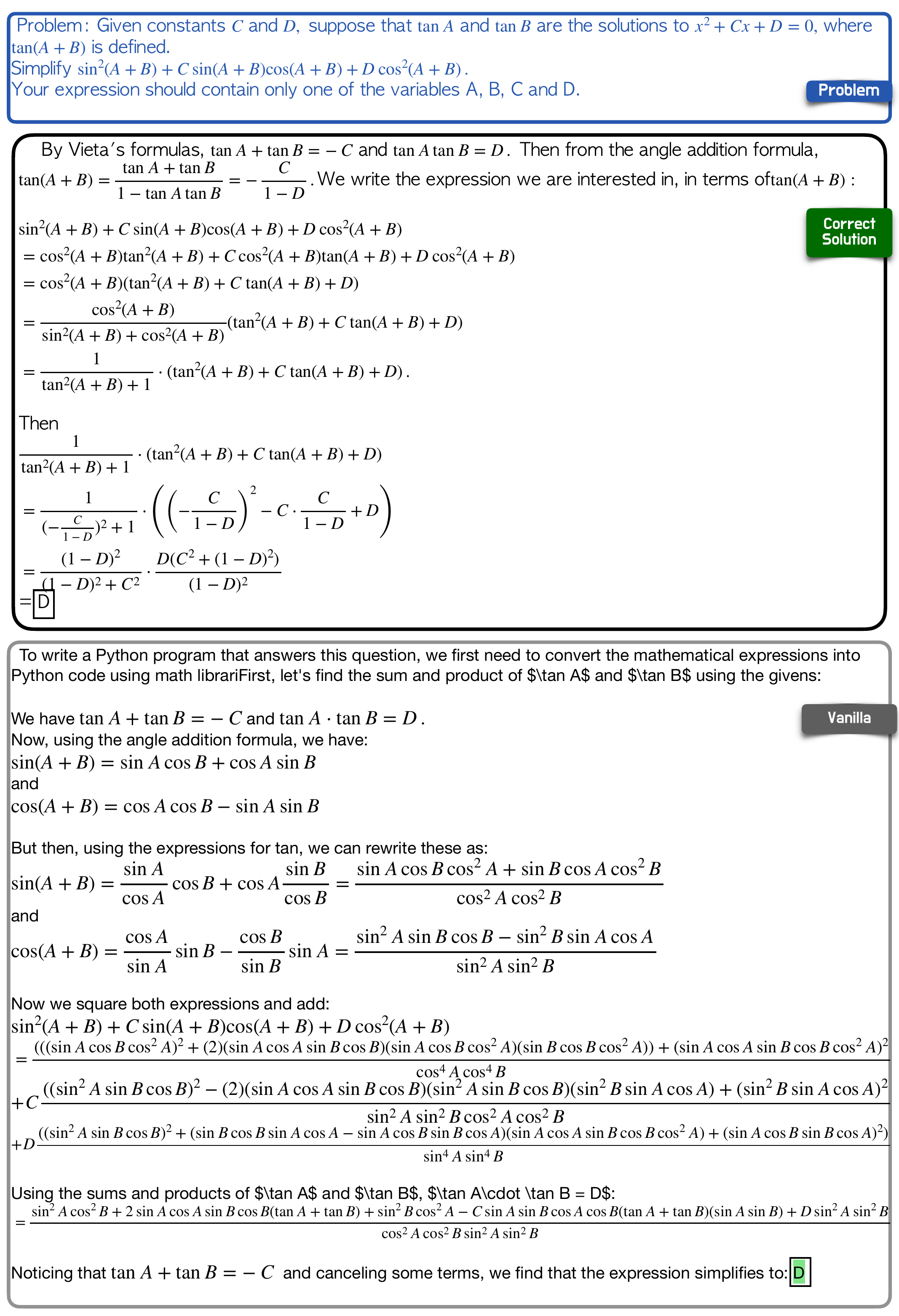}
\caption{The precalculus problem where other methods are correct but \MathChat is wrong. This figure shows the ground truth solution and the response with vanilla prompt.}
\label{app:precal_1}
\end{figure}

\begin{figure}[tb]
\centering
\includegraphics[width = 1\hsize]{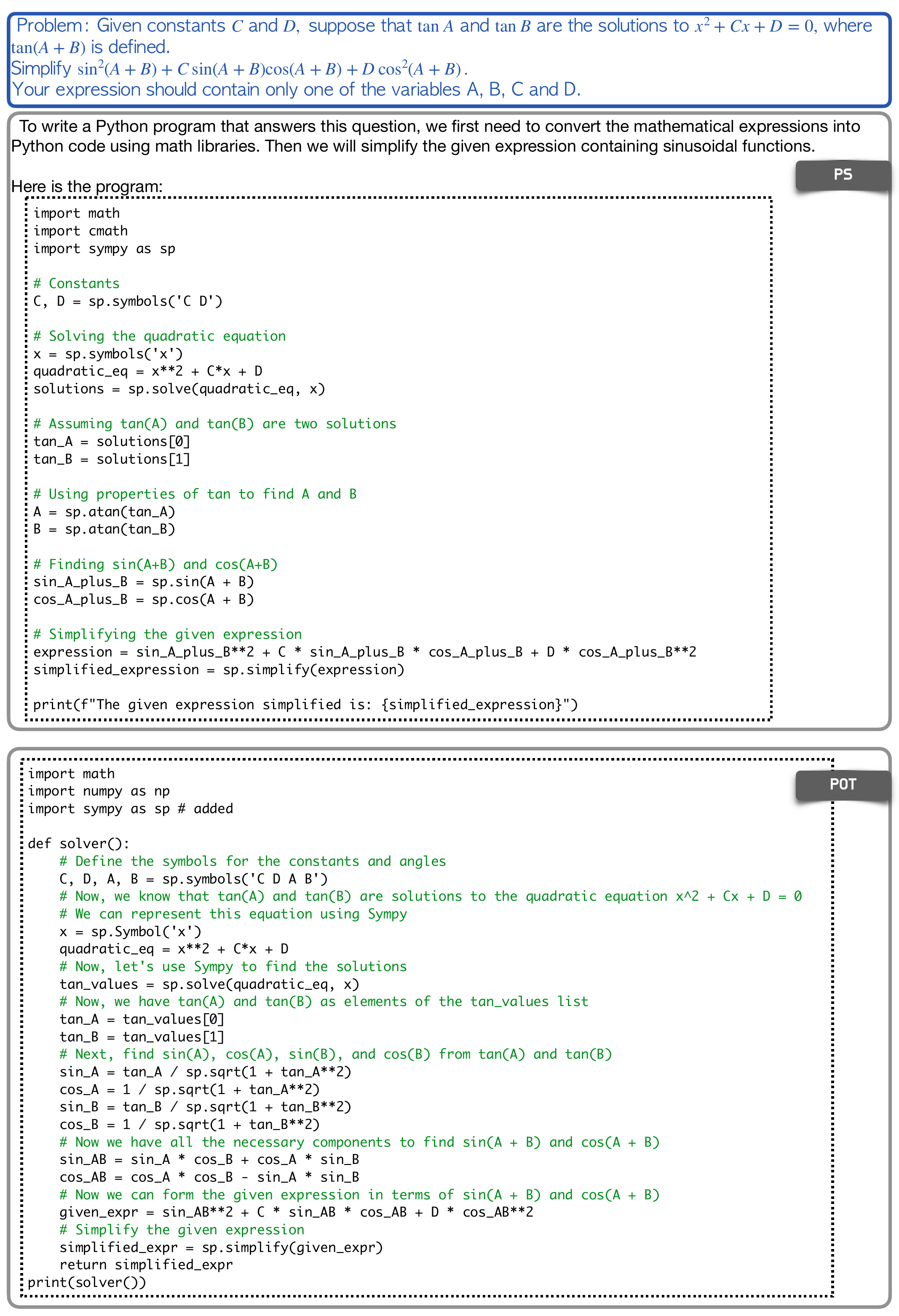}
\caption{The precalculus problem where other methods are correct but \MathChat is wrong (Continued). This figure shows the PS and PoT code. Both code returns the correct result: \texttt{"D"}.}
\label{app:precal_2}
\end{figure}

We present the precalculus example where \MathChat fails but all other methods success (Figure~\ref{app:precal_1}, Figure~\ref{app:precal_2}, Figure~\ref{app:precal_3}). The results from PS and PoT show that it is easy to get this problem correct with Python (using the \texttt{sympy.simplify} function).
However, in \MathChat, the LLM agent chooses to solve the problem via direct reasoning. Both \MathChat and vanilla prompt solve this problem by writing extremely long derivations. \MathChat solves the problem with an even longer step-by-step response and makes a calculation error during the process.

Additionally, we also provide an overview of the number of problems where all methods either fail or succeed in Table~\ref{overall_result}. 

\begin{table}[htb!]
\centering
\begin{tabular}{@{}c|cccccc|c@{}}
\toprule[1.5pt]
& Algebra & C.Prob & I.Alg & N.Theory & Prealg & Precalc & Total\\
\midrule

All Success          & 46 &  13  & 0  & 18 & 45 & 1 & 176 \\
All Fail          & 57 &  32  & 171  & 20 & 36 & 86 & 402 \\

\bottomrule[1.5pt]
\end{tabular}
\vspace{10pt}
\caption{The number of problems where all methods fail, and all methods succeed.}
\label{overall_result}
\end{table}

\subsection{The Relationship Between Failure Rate and Generated Solution Length}

\begin{figure}[!tb]
\centering
\includegraphics[width = 0.49\hsize]{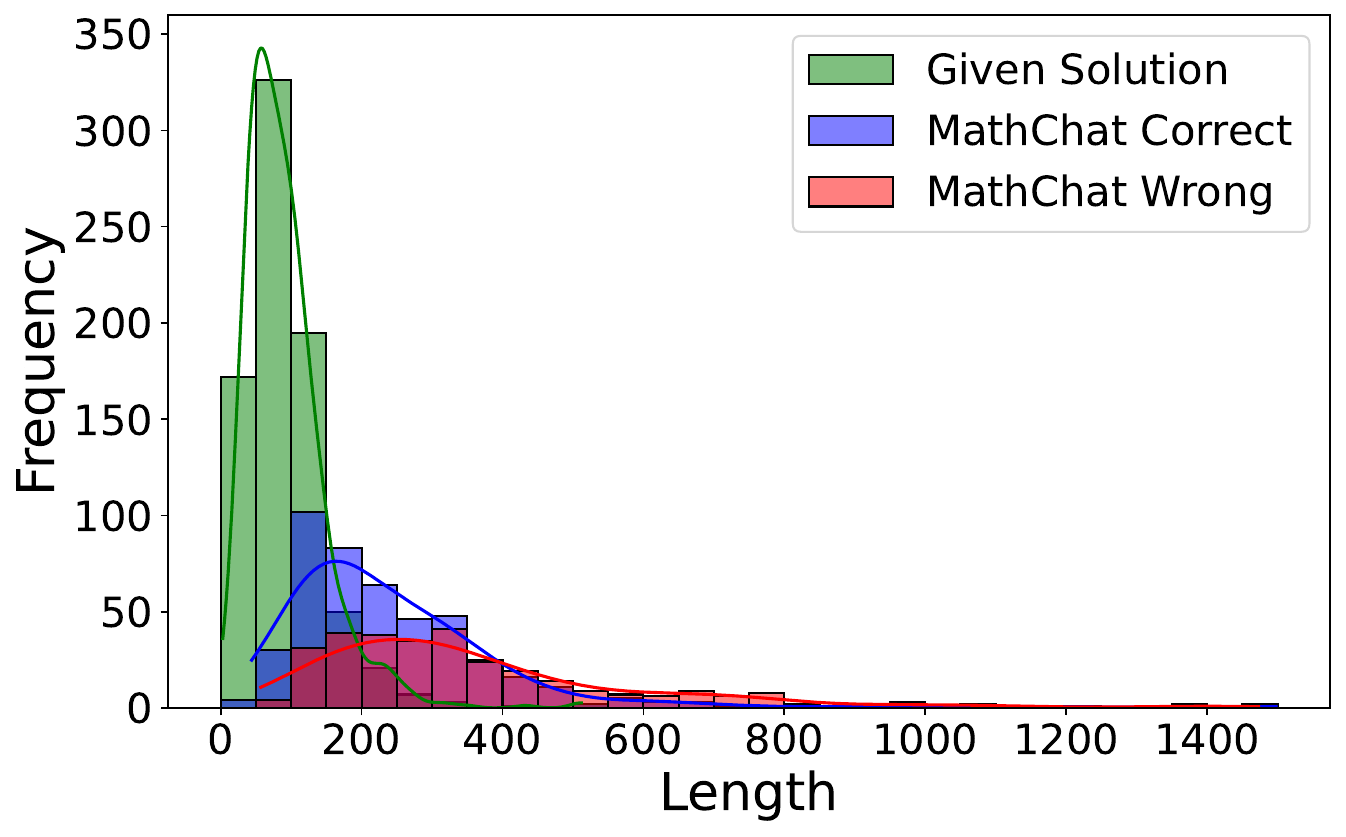}
\includegraphics[width = 0.48\hsize]{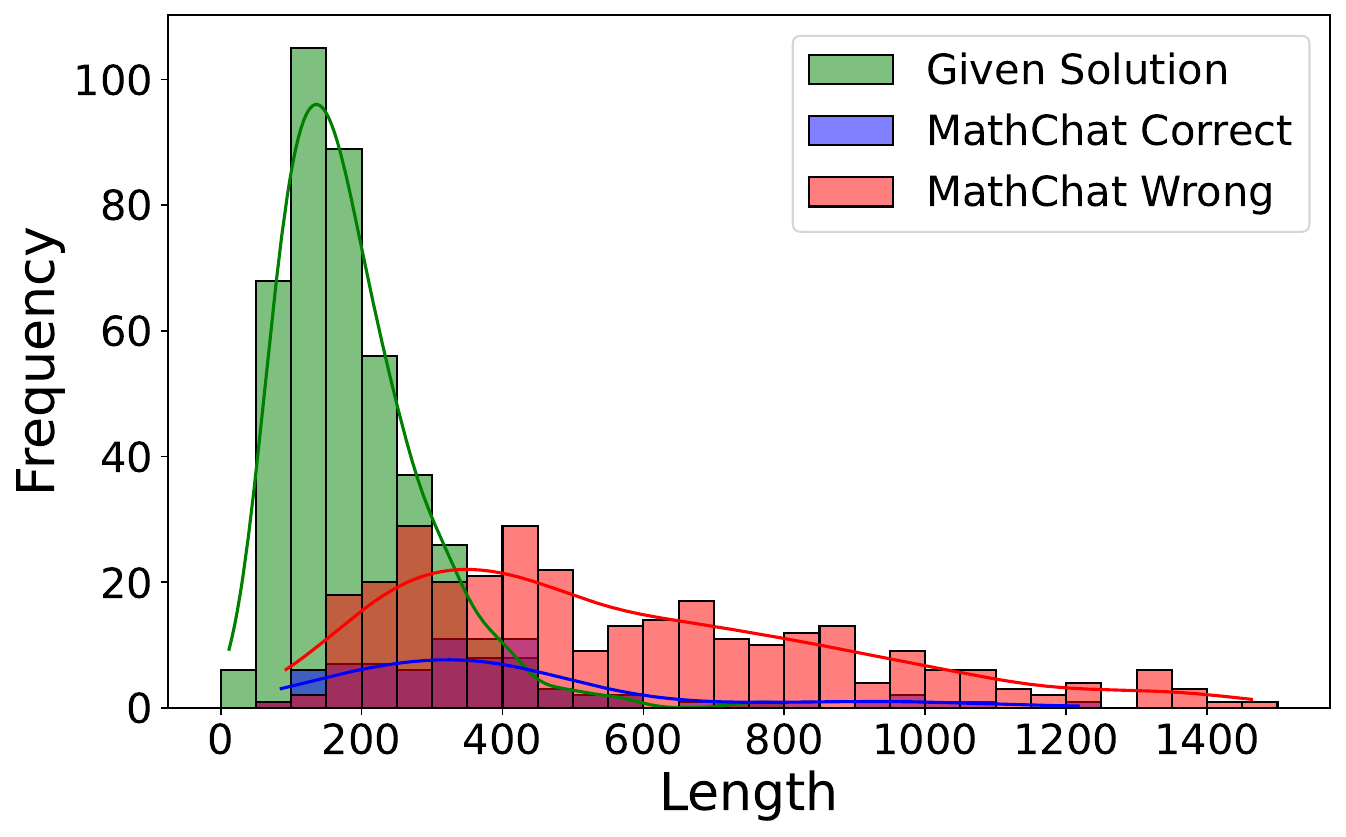}
\caption{Distribution of solution length of both correctly and incorrectly solved problems in \MathChat. The distribution of length of the given solution (ground truth) is also shown. The left figure represents the distribution of the less challenging categories and the right figure represents problems from Intermediate Algebra and Precalculus. We cut off outliers that the split string length is longer than 1500.}
\label{app:length_dist}
\end{figure}

Chain of Thought (CoT) prompting shows that extra reasoning steps for a problem can improve the ability of LLMs~\cite{wei2022chain}. With GPT-4, explicit reasoning is no longer an issue. Instead, we find that a long and tedious reasoning process may result in more type 2 failures, such as calculation errors, which results in a wrong answer even the overall direction is correct. We plot the distribution of correct and wrong answer lengths and also the answer length of the given solution (The length of the string list from splitting with a single space is used here). Since more complex and challenging problems are likely to have a longer solving process but still a lower success rate, we separate problems from Intermediate Algebra and Precalculus with other categories (Figure~\ref{app:length_dist}), to distinguish less challenging problems from harder problems. We note that the success rate of \MathChat on the four less challenging categories goes over 50\%, but the rate is lower than 20\% for Intermediate Algebra and Precalculus.

\begin{figure}[tb]
\centering
\includegraphics[width = 1\hsize]{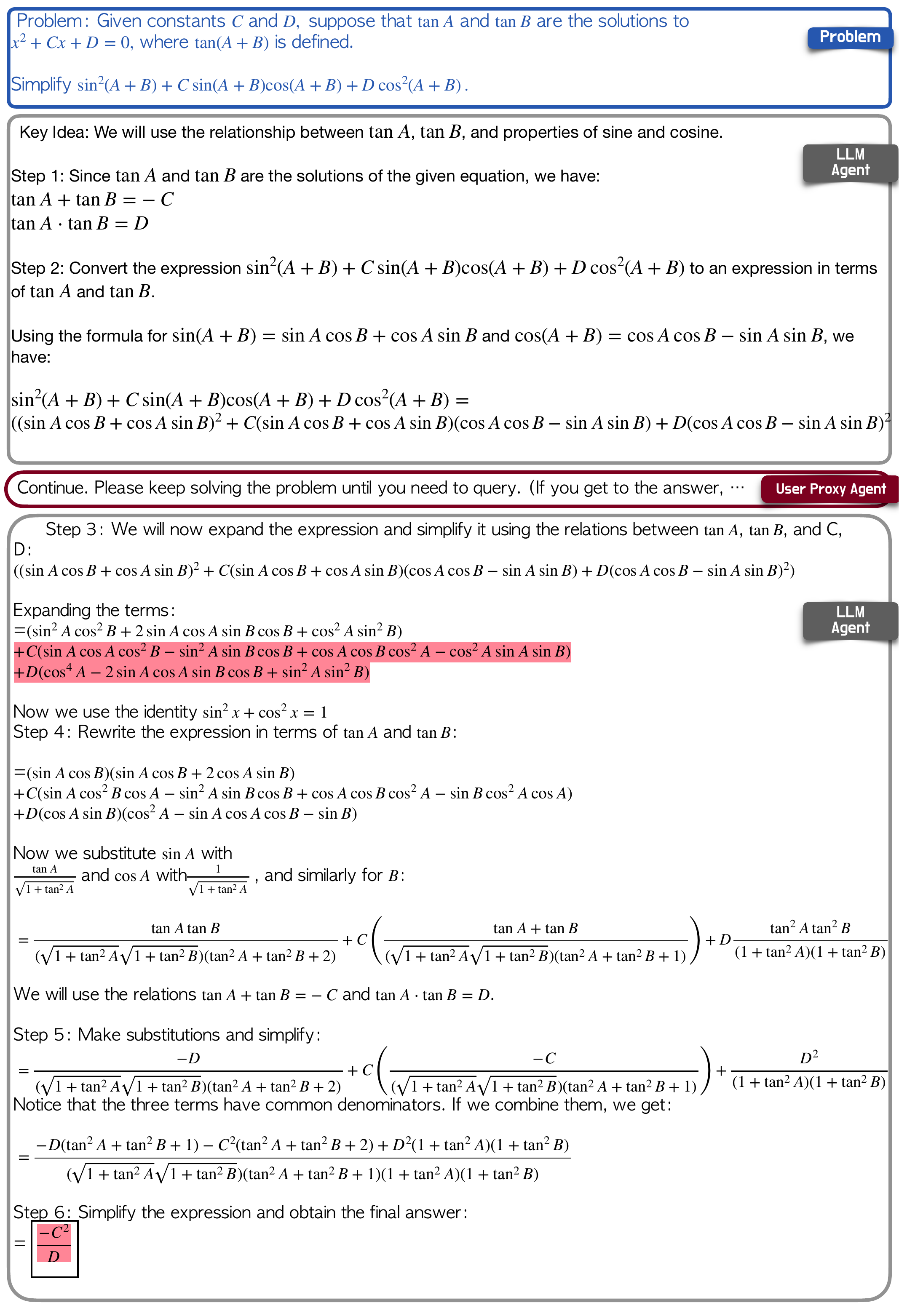}
\caption{The precalculus example where all the other methods are correct but \MathChat is wrong (Continued). This figure shows the conversation generated in \MathChat. The LLM agent in \MathChat chooses to solve the problem via direct reasoning, and it makes a calculation error when expanding the terms in Step 3.}
\label{app:precal_3}
\end{figure}

Overall, the solution length of \MathChat is longer than the ground truth solution. The length of the given solution on the two fundamentally challenging categories is longer than other categories. For \MathChat, correct answers and wrong answers from the less challenging categories have a similar distribution in solution length, where the majority of problems are solved with 50 to 500 string length. For harder problems, however, an answer with more than 600 string lengths is likely to be wrong. From the precalculus problem shown in Figure~\ref{app:precal_3}, the LLM agent can choose a plausible strategy to solve the problem, but that strategy is less efficient and involve more math operations compared to the given solution, this results in a much longer response, and it is more likely to make errors during the process.

\end{document}